\documentclass{article}

\usepackage{mathtools}

\usepackage{amsmath}
\usepackage{amssymb}
\usepackage{amsthm}

\usepackage{microtype}
\usepackage{graphicx}
\usepackage{subcaption}

\usepackage{hyperref}

\usepackage[preprint]{ICML2026/icml2026}

\usepackage[capitalize,noabbrev]{cleveref}

\theoremstyle{plain}

\theoremstyle{definition}

\theoremstyle{remark}

\usepackage{algorithm}

\usepackage{hyperref}
\usepackage{url}

\usepackage[utf8]{inputenc} % allow utf-8 input
\usepackage[T1]{fontenc}    % use 8-bit T1 fonts
\usepackage{hyperref}       % hyperlinks
\usepackage{url}            % simple URL typesetting
\usepackage{booktabs}       % professional-quality tables
\usepackage{amsfonts}       % blackboard math symbols
\usepackage{nicefrac}       % compact symbols for 1/2, etc.
\usepackage{microtype}      % microtypography
\usepackage{xcolor}         % colors
\usepackage{comment}

\usepackage{amsthm}
\usepackage{acronym}
\usepackage{array}
\usepackage{wrapfig}
\usepackage{multirow}
\usepackage{tabu}
\usepackage{tcolorbox}
\usepackage{enumitem}
\usepackage{graphicx}
\usepackage{caption}
\usepackage{float}
\usepackage{subcaption}

\usepackage{listings}
\usepackage{color}

\definecolor{codegreen}{rgb}{0,0.6,0}
\definecolor{codegray}{rgb}{0.5,0.5,0.5}
\definecolor{codepurple}{rgb}{0.58,0,0.82}
\definecolor{backcolour}{rgb}{0.95,0.95,0.92}
\definecolor{darkgreen}{rgb}{0,0.5,0}

\usepackage[colorinlistoftodos,textsize=tiny]{todonotes}
\usepackage{pifont}

\usepackage{enumitem}

\usepackage{algorithm}
\usepackage{algorithmic}  % or \usepackage{algpseudocode}

\newcommand{\Voc}{\ensuremath{\mathcal{V}}}

\newcommand{\X}{\ensuremath{\mathbf{x}}}
\newcommand{\Z}{\ensuremath{\mathbf{z}}}
\newcommand{\Y}{\ensuremath{\mathbf{y}}}
\newcommand{\ft}{\ensuremath{\tilde{f}}}
\newcommand{\Xt}{\ensuremath{\tilde{\mathbf{x}}}}
\newcommand{\Yt}{\ensuremath{\tilde{\mathbf{y}}}}

\newcommand{\V}{\ensuremath{\mathcal{V}}}
\newcommand{\Pd}{\ensuremath{\mathcal{P}}}

\usepackage{calc}

\usepackage{longtable}
\usepackage{xcolor}

\usepackage{adjustbox}

\icmltitlerunning{
Language Model Inversion through End-to-End Differentiation
}

\begin{document}

\twocolumn[
\icmltitle{
%On the Invertibility of Language Models via End-to-End Differentiable Language Models
%Language Model Inversion via Backward Pass
Language Model Inversion through End-to-End Differentiation
}



  \icmlsetsymbol{equal}{*}

  \begin{icmlauthorlist}
    \icmlauthor{Kevin Yandoka Denamganaï}{yyy}%{equal,yyy}
    \icmlauthor{Kartic Subr}{yyy}%{equal,yyy,comp}
    % \icmlauthor{Firstname3 Lastname3}{comp}
  \end{icmlauthorlist}

  \icmlaffiliation{yyy}{School of Informatics, University of Edinburgh, Edinburgh, UK}
  %\icmlaffiliation{comp}{Company Name, Location, Country}
  %\icmlaffiliation{sch}{School of ZZZ, Institute of WWW, Location, Country}

  \icmlcorrespondingauthor{Kevin Yandoka Denamganaï}{kevin.denamganai@ed.ac.uk}
  %\icmlcorrespondingauthor{Firstname2 Lastname2}{first2.last2@www.uk}

  \icmlkeywords{Machine Learning, ICML}

  \vskip 0.3in
]



\printAffiliationsAndNotice{}  % no special notice (required even if empty)

\begin{abstract}
Despite emerging research on Language Models (LM), few approaches analyse the invertibility of LMs. That is, given a LM and a desirable target output sequence of tokens, determining what input prompts would yield the target output remains an open problem. We formulate this problem as a classical gradient-based optimisation. First, we propose a simple algorithm to achieve end-to-end differentiability of a given (frozen) LM and then find optimised prompts via gradient descent. Our central insight is to view LMs as functions operating on sequences of distributions over tokens (rather than the traditional view as functions on sequences of tokens). 
Our experiments and ablations demonstrate that our DLM-powered inversion can reliably and efficiently optimise prompts of lengths $10$ and $80$ for targets of length $20$, for several white-box LMs (out-of-the-box). 
%\KS{up to 80?}, 
\end{abstract}

\section{Introduction}

We consider the problem of inverting a Language Model (LM), with the goal of hypothesising a prompt (input sequence of tokens) that, under a frozen pretrained LM, would yield some given \emph{target output} sequence of tokens. Indeed, the \emph{recovered prompt} is unlikely to be unique, and so we formulate the problem as a gradient-based optimisation to determine the `best' input prompt, based on a given distance metric between target outputs. We present a simple method to back-propagate gradients through LMs. 

LMs are composed of a tokenizer, an embedding module, a core module (Transformer~\citep{vaswani2017attention}, Mamba~\citep{dao2024transformers}, etc) and a next-token (categorical) sampling module (see Figure~\ref{fig:DLMI} a.). Of these,  the embedding module and the next-token sampling module are the two non-differentiable modules, preventing gradients to be back-propagated end-to-end through LMs. 

First, we use a soft embedding module that reformulates the lookup operation (hard embedding) by modelling the input as a distribution over tokens (rather than as sampled tokens). This requires a shift in perspective similar to that hinted at by soft prompting literature~\citep{li2021prefix-tuning, chang2024efficient-prompting-survey}, of LMs as functions from and to distributions over tokens (SDoTs) rather than over the space of tokens (SoTs) as explained in  Section~\ref{sec:distributional-shift}.  

To address the challenge of differentiating through a categorical sampling operation (such as in next-token sampling), several  methods are available but they have not been used in this context within LMs. For example, REINFORCE ~\citep{williams1992simple}, the Concrete distribution~\citep{maddison2017concrete}, the Gumbel-Softmax (GS) gradient estimator~\citep{jang2017categorical}, etc. We adopt GS gradient estimations in conjunction with decoupled temperature learning~\citep{bengio2013estimating,Havrylov2017}.
Our contributions are:
\begin{enumerate}[topsep=0em,itemsep=-.5em, leftmargin=1.2em]
    \item our formulation achieves end-to-end differentiable LMs;
    \item our simple algorithm optimises prompts to produce target outputs for any white-box LM; and
    \item our method effectively optimises prompts yielding target sequences that are both fluent as well as having high perplexity under the subject LM.
\end{enumerate}

\section{Related Works \& Background}
\label{sec:related-works}
\label{sec:background}
% \KD{I have the impression that  I need to provide a lot of background already to discuss the nuances with those related works. Thus, shouldn't we put related works at the end of the paper in a form of `Discussion \& Related Works' section?}

\subsection{Language Model Inversion \& Current Limitations}
\label{subsec:language-model-inversion}

The LMI problem consists of finding a fully-parameterised prompt $\X\in\V^*$ to a given target output $\Y\in\V^*$ for a given subject LM $f_\theta$ over vocabulary $\V$ and with parameters $\theta$. 
However, previous literature (cf. Table~\ref{table:lit-review-LM-inversion}) has focused on an evaluation protocol where inverters have to reconstruct ground-truth prompt $\bar{\X}$ from target output text $\Y$ (or logits) of a subject LM, where the ground-truth $(\bar{\X}, \Y)$ pair comes from Natural Language / fluent datasets~\citep{morris2023Vec2text,morris2024LMI,zhang-etal-2024-extracting-O2P,li-klabjan-2025-RPE}.

Those previous works report measures akin to Exact Match (EM) and/or cosine similarity between reconstructed prompt $\X$ and ground-truth prompt $\bar{\X}$, respectively, at the token or the embedding level. 
This evaluation protocol bears 2 major issues.
Firstly, it lacks internal validity, because it implicitly assumes that there is only a single prompt $\X\in\V^*$ that would elicit a given target output $\Y\in\V^*$, 
%effectively assuming an inverse function, 
whereas the combinatorial nature of the language modelling problem would suggests 
%a right inverse relation rather, 
a plurality of prompts to any given target output. 
Thus, when the retrieved prompt $\X$ is different from the ground-truth prompt $\X$ but still eliciting the target output $\Y$ as a completion via the subject LM, the metric counts it as a failure, and thus this evaluation protocol results in inflated false negatives. 
Secondly, the reliance of the evaluation protocol on fluent datasets means that evaluation only covers a very small and controlled subset of possible target outputs of the subject LM, which are likely to be in-distribution to any given pre-trained, subject LM. The thus-measured performance of the inverter is therefore lacking external validity: it is unclear how would the inverter perform when deployed on variously-fluent target outputs. 
%Thirdly, it is not aligned with our objective of eliciting specific behaviours/outputs of any kind, as it focuses on instruction-following datasets only.

The contemporary and state-of-the-art work of ~\citet{skapars2025SODA-gpt}, proposing the \textbf{Sparse One-hot Discrete Adam (SODA)} inverter algorithm, includes evaluation over fluent and non-fluent (because randomly-generated) target outputs. 
% We acknowledge it as a first step in the right direction to increase the external validity of the evaluation, and
We further parameterise the fluency of the evaluation target outputs with respect to the subject LM that is being studied (cf. Section~\ref{sec:dataset-generation}), enabling us to better appreciate the generalisation abilities of the evaluated inverters.

% \item{
% The EM measures in the context of prompt reconstruction is also an imperfect metric, because it assumes that there is only one single antecedent prompt that can forward map to the target output provided as input to the approach. However, from the best of our knowledge, it is unclear whether LMs are injective, and it is rather more sound to assume that they are surjective, given that Natural Languages are combinatorial in nature.
% }
Regarding the internal validity issue, we argue that it is necessary to report EM-like measures, not between prompts, but between specified target outputs $\Y$ and $\hat{\Y} = f_\theta(\X)$ which is completion of the subject LM after conditioning on the optimised prompts $\X$ is necessary. 
This evaluation over text has only been seen in the margin of LMI literature, to the best of our knowledge, in the works of ~\citet{jones2023-automatically-auditing-LM-ICML2023} and ~\citet{zou2023GCG}.
%(as well as in ~\citet{PEZ} in the context of visual outputs). 
However, these two works still only considered fluent target outputs. 

In the context of adversarial attacks, i.e. searching for \textit{adversarial examples}~\citep{biggio2013evasion,szegedy2013intriguing}, which are prompts that encourages prediction errors in LMs,~\citet{guo-etal-2021-GBDA-EMNLP2021} proposed \textbf{Gradient-Based Distributional Attack (GBDA)}. 
%While it is a strong baseline for adversarial attacks, it is inefficient at the \textit{reverse large language model} task from ~\citet{jones2023-automatically-auditing-LM-ICML2023}, and our results show that it is also inefficient in the general case of the LMI problem (cf. Section~\ref{sec:invertibility-evaluation}). 
While it is a strong baseline for adversarial attacks, it has been found to be inefficient in the context of LMI~\citet{jones2023-automatically-auditing-LM-ICML2023}.  
Rather than LMI, their actual focus is to systematically investigate LMs for possible bias, error, security flaw, or misalignment to ensure that they are safe, reliable, and ethical. 
Consequently, their studies are limited to LMI for fluent target outputs of length up to $M=3$ and prompt length up to $N=8$. In contrast, we study LMI problem over a spectrum of fluency of the target output with lengths of $M=20$ and various prompt length $N\in[10,80]$ (cf. Section~\ref{sec:invertibility-evaluation}), which amount to one order of magnitude longer lengths. 

Our results also show that GBDA is inefficient in our more general investigation of the LMI problem. 
Nevertheless, like our work, \textbf{GBDA} relies on both a Gumbel-Softmax (GS) gradient estimator over the optimised input prompt and a soft embedding module within the LM autoregressive pipeline (cf. Table~\ref{tab:dlmi_ablations}). 
It lacks the critical element of jointly learning the optimised prompt and the (decoupled) temperature(s) $\tau$ of the tempered-softmax in the GS trick. 
%(as well as the teacher forcing trick which improves results). 
Thus, we acknowledge that the work of ~\citet{guo-etal-2021-GBDA-EMNLP2021} is the first work to contain a \textbf{theoretically-sound} end-to-end DLM extension, in hindsight and without ever them claiming it. 
However, we argue that our work is the first to propose a \textbf{practically-efficient \& theoretically-sound} end-to-end DLM extension, which makes the LMI problem solvable at reasonable scales (both in terms of prompt and target lengths and LM parameter size), in an efficient and reliable manner (as it is based on an any-time algorithm).

\subsection{Why are LMs not end-to-end differentiable?}

The inability to perform end-to-end differentiation through LMs stems from two SoT-processing, non-differentiable components inherent to the autoregressive pipeline relying on SoT as the main abstraction. 
%: the embedding module that maps discrete tokens to continuous representations, and the sampling module that converts continuous logits back to discrete token selections. 
%Note the reliance on SoT as the main abstraction of this pipeline. 
% In the following, we examine the foundations of this paradigm.

% \paragraph{Transformer Architecture and the Autoregressive Pipeline.} 
% The modern LM paradigm was established by the Transformer architecture~\citep{vaswani2017attention}, which introduced the self-attention mechanism as the primary computational primitive for sequence processing. 
% Whether deployed in encoder-only, decoder-only, or encoder-decoder configurations, Transformers have codified the autoregressive generation pipeline that dominates contemporary Natural Language Processing. 
% %Thus, LMs generate text sequentially, conditioning each token prediction on all previously generated tokens through causal masking. 
% While the internal computations of the Transformer module ---comprising multi-head self-attention layers, feed-forward networks, and layer normalization of different flavours---remain fully differentiable, the LM pipeline's input and output interfaces introduce fundamental discontinuities that prevent gradient to flow end-to-end through the complete generation process.

\begin{figure*}[t]
\centering
\includegraphics[width=0.95\linewidth]{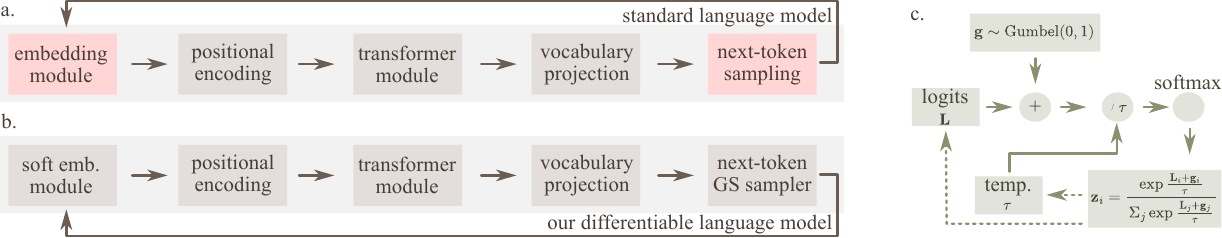}
\caption{\label{fig:DLMI} Our abstracted view of (a) existing LMs; (b) our differentiable LM and (c)  gradients through sampling~\citep{bengio2013estimating}. }
\end{figure*}

\paragraph{The Two Non-Differentiable Barriers.} 
The (hard) embedding module is the first barrier to differentiability. 
This component performs a lookup operation that maps discrete token indices 
%from a finite vocabulary $\V$, over which the LM is defined, to dense, continuous vector representations in $\mathbb{R}^d$, where $d$ is the embedding space dimensionality. 
to dense, continuous vector representations. 
Formally, given a vocabulary $\V$ of size $V=|\V|$ and embedding dimension $d$, the embedding function $E: \{1, 2, \ldots, V\} \rightarrow \mathbb{R}^d$ is implemented as:
$
\mathbf{e}_i = E(i) = \mathbf{W}_E[i]
$ 
where $\mathbf{W}_E \in \mathbb{R}^{V \times d}$ is the embedding matrix and the operation extracts the $i$-th row corresponding to token index $i$. 
This lookup operation is non-differentiable with respect to the token index $i$ itself: while gradients can flow backward through the embedding vectors to update $\mathbf{W}_E$, there exists no meaningful gradient with respect to the discrete selection. The derivative $\frac{\partial E(i)}{\partial i}$ is undefined because the domain of $E$ is discrete.

%The historical development of embedding techniques provides important context for understanding this architectural choice. 
% Early neural LMs employed distributed word representations, with seminal work including Word2Vec~\citep{mikolov2013word2vec}, which introduced the continuous bag-of-words (CBOW) and skip-gram architectures for learning word embeddings from unlabeled text, and GloVe (Global Vectors for Word Representation)~\citep{pennington2014glove}, which derived embeddings from word co-occurrence statistics in large corpora. 
% These foundational methods established the paradigm of representing discrete linguistic units as continuous vectors that can subsequently be processed by neural networks, but it introduced the fundamental discretization that precludes end-to-end differentiation.

The second barrier manifests at the output stage of the pipeline through the next-token sampling module (cf. Figure~\ref{fig:DLMI}). 
After a $t$-long input SoT $\X_{<t}\in\V^t$ has been transformed by the embedding module and subsequently processed by core modules over the embedding space, `vocabulary projection' produces a distribution over the vocabulary by computing logits $\mathbf{z} \in \mathbb{R}^V$ and applying the softmax function to obtain probabilities for the next-token $\X_t\in\V$:
\begin{equation}
p(\X_t = i \mid \X_{<t}) = \frac{\exp(z_i / \tau)}{\sum_{j=1}^{V} \exp(z_j / \tau)}
\end{equation}
where $\tau$ is a temperature parameter controlling the sharpness of the distribution. The actually-problematic token selection, however, requires sampling from this distribution (or performing deterministic selection via argmax), which introduces another discrete operation that is not differentiable with respect to the logits $z$, inputs of the next-token sampling module. 
%The argmax function has zero gradients almost everywhere and undefined gradients at tie points, while sampling introduces stochasticity that fundamentally breaks the deterministic gradient computation required for backpropagation.
%\textbf{The Critical Importance of the Next-Token Sampling Module.} 
Note that the nature of the next-token sampling module carries profound implications for model performance, extending far beyond mere implementation detail as it is can be the difference between an LM employing Chain-of-Thoughts~\citep{wei2022chain} reasonings or not~\citep{wang2024-CoTWithoutPrompting}. 
%This importance has become particularly salient with recent advances in the context of LM's post-training with Reinforcement Learning from Verifiable Rewards (RLVR)~\citep{guo2025deepseek}. Indeed, these models demonstrate that the choice of sampling scheme directly determines both the diversity and accuracy of generated reasoning chains. 
Reinforcement Learning from Verifiable Rewards (RLVR) post-training approaches have exacerbated the matter, as the exploration-vs-exploitation trade-off inherent to this post-training technique requires carefully calibrated sampling approaches, typically involving temperature-adjusted sampling or nucleus (top-$p$) sampling, to effectively explore the space of potential solution paths while maintaining coherence and correctness.

\subsection{Gradients through categorical distributions}
\label{subsec:gradient-estimators}

The fundamental non-differentiability of the next-token sampling module in the LM pipeline necessitates the development of gradient estimation techniques that enable backpropagation through stochastic computational nodes~\citep{bengio2013estimating}. %This section recalls the gradient estimation problem in the context of stochastic computational nodes, the bias-variance trade-offs inherent to different estimators, and the theoretical foundation for the Gumbel-Softmax family of methods that have emerged as practical solutions for enabling end-to-end learning through categorical distributions.
%\paragraph{Differentiation through Stochastic Computational Nodes.} 
Consider a computational graph containing a stochastic node that samples a discrete random variable $z \sim p_\theta(z)$ from a categorical distribution parameterized by $\theta$. 
Let $f(z)$ denote a downstream function that depends on this discrete sample, with the overall objective being to minimize an expected loss $\mathcal{L}(\theta) = \mathbb{E}_{z \sim p_\theta(z)}[f(z)]$. 
Following ~\citet{bengio2013estimating}, we seek to estimate $\nabla_\theta \mathcal{L}(\theta)$ despite the discrete nature of $z$ preventing direct differentiation through the sampling operation.
%:
%\begin{equation}
%\nabla_\theta \mathcal{L}(\theta) = \nabla_\theta \mathbb{E}_{z \sim p_\theta(z)}[f(z)] = \mathbb{E}_{z \sim p_\theta(z)}[\nabla_\theta \log p_\theta(z) \cdot f(z)]
%\end{equation}
%where the equality follows from the log-derivative trick (also known as the REINFORCE trick~\citep{williams1992simple} or score function estimator). 
The challenge lies in constructing practical estimators $\hat{g}(\theta)$ that approximate this gradient using finite samples while maintaining desirable statistical properties, such as low variance and bias. 
We direct the reader to Appendix~\ref{sec:further-background} for further details on gradient estimators, from the common REINFORCE~\citep{williams1992simple} to the more advanced REBAR~\citep{tucker2017rebar} and their properties. 
In this work, we will focus mainly on the Gumbel-Softmax (GS) gradient estimator.

\paragraph{Gumbel-Softmax and Concrete Distributions: A Biased Low-Variance Gradient Estimator.} 
The high variance of REINFORCE motivated further developments focusing on continuous relaxations of categorical distributions. 
Two independently proposed methods---the Gumbel-Softmax (GS) distribution~\citep{jang2017categorical} and the Concrete distribution~\citep{maddison2017concrete}---provide mathematically equivalent formulations of this approach. Both methods introduce a continuous approximation to discrete sampling that permits backpropagation through the sampling operation and therefore enable the estimation of gradients.

The GS distribution constructs a continuous relaxation by adding Gumbel noise to logits and applying a temperature-scaled softmax. 
Given a categorical distribution with 
%class probabilities $\pi_1, \ldots, \pi_K$ (where $\log \pi_k$ represents the unnormalized logits), 
unnormalized logits $z_1, \ldots, z_K$, the GS sample is:
\begin{equation}
%y_k = \frac{\exp((\log \pi_k + g_k) / \tau)}{\sum_{j=1}^{K} \exp((\log \pi_j + g_j) / \tau)}
y_k = \frac{\exp((z_k + g_k) / \tau)}{\sum_{j=1}^{K} \exp((z_j + g_j) / \tau)}
\end{equation}
where $g_k \sim \text{Gumbel}(0, 1)$ are independent Gumbel random variables and $\tau > 0$ is a temperature of the tempered softmax. 
The resulting vector $\mathbf{y} = (y_1, \ldots, y_K)$ lies on the probability simplex and can be used as a soft, differentiable approximation to a one-hot encoded discrete sample from a categorical distribution.

Gradients are estimated by treating $\mathbf{y}$ as a deterministic function of the Gumbel noise and the parameters, enabling backpropagation through the reparameterization:
\begin{equation}
\hat{g}_{\text{GS}}(\theta) = \nabla_\theta f(\mathbf{y}(\theta, \mathbf{g}))
\end{equation}
%This estimator exhibits lower variance than REINFORCE, as it leverages the smoothness of $f$ with respect to continuous inputs. 
While having lower variance than REINFORCE, the GS estimator is biased because $\mathbf{y}$ is not a true discrete sample but rather a continuous relaxation, introducing systematic error into the gradient estimates.

\paragraph{The Temperature Trade-off: Bias versus Variance.} 
The temperature parameter $\tau$ in Gumbel-Softmax-based estimators governs a fundamental bias-variance trade-off. 
As $\tau \to 0$, the GS samples appear to be drawn from a distribution that approaches the true categorical distribution defined by logits $z_1, \ldots, z_K$, thus reducing bias but increasing variance as the gradients become increasingly concentrated on a single mode. 
Conversely, as $\tau \to \infty$, the GS samples appear to be drawn from a distribution that becomes more uniform, reducing variance but increasing bias as the continuous relaxation diverges further from the true discrete distribution.

\paragraph{The Importance of Learned Temperature Parameters.} 
%A critical practical refinement to Gumbel-Softmax-based methods involves learning the temperature parameter $\tau$ jointly with the model parameters rather than treating it as a fixed hyperparameter. 
In the context of Emergent Communication research~\citep{lazaridou2020emergent}, ~\citet{Havrylov2017} introduced the practical refinement of learning the temperature parameter $\tau$ jointly with the model parameters rather than treating it as a fixed hyperparameter of the GS-based gradient estimator. 
It substantially improves sample efficiency during training. 
%Rather than manually scheduling temperature annealing or performing extensive hyperparameter searches, learned temperatures enable the optimization process to automatically navigate the bias-variance trade-off, adapting the temperature to the current stage of learning and the specific characteristics of the task.
Subsequent work has built upon this, investigating the impact of the technique on the downstream considerations~\citep{DenamganaiAndWalker2020b} . The success of learned temperatures in Emergent communication suggests their potential value for LMI and other applications requiring gradient estimation through categorical distributions. %By incorporating temperature as a learnable parameter, optimization can dynamically adjust the balance between gradient accuracy and stability, potentially accelerating convergence and improving final performance without requiring manual intervention or domain-specific tuning strategies.

{\scriptsize
\begin{algorithm}[ht]
\caption{DLMI}
\label{alg:dlmi}
\begin{algorithmic}[1]
\REQUIRE 
$f_\theta$, 
$\mathbf{y}$, 
$\Voc$, 
$T_\text{opt}$, 
$N$, 
$\tau_0$, 
$d(\cdot, \cdot)$,  
% (e.g. CrossEntropy or any proxy to the distance metric $d$ of the LMI problem)
$\epsilon>0$ \\
% Frozen LM 
% target 
% vocabulary 
% max number of optimization steps 
% inverse prompt length 
% initial temperature 
% loss function 
%  to guarantee numerical stability,
\ENSURE Optimized logits $\mathbf{Z}^\ast$
\vspace{1em}
\STATE $\mathbf{Z} \leftarrow \mathcal{N}(0, 1)^{N \times |\Voc|}$ \COMMENT{Random initialization}
\STATE $\Phi \leftarrow \mathcal{N}(0, 1)^N$ \COMMENT{Random initialization}
\STATE $\mathbf{E} \leftarrow$ embedding matrix over $\Voc$
\FOR{$t = 1$ to $T_\text{opt}$}
    
    \STATE [] \textcolor{gray}{\# GS Next-Token Sampling from logits $Z$:}
    \STATE $\mathbf{g} \sim \text{Gumbel}(0,1)^{n \times |\Voc|}$ 
    % \\ \COMMENT{$g_i = -\log(-\log(u_i))$, $u_i \sim \text{Uniform}(\epsilon, 1-\epsilon)$}
    \STATE $\tau \leftarrow \epsilon + \tau_0 (1+\tanh(\Phi))$
    \STATE $\mathbf{p}_{\text{soft}} \leftarrow \text{softmax}\left((\mathbf{Z} + \mathbf{g}) / \tau\right)$ 
    
    %\STATE \textcolor{gray}{\# Compute soft embeddings and forward pass}
    %\STATE \textcolor{gray}{\# Compute soft embeddings:}
    %\STATE $\mathbf{h} \leftarrow \mathbf{p}_{\text{soft}} \cdot \mathbf{E}$ %\COMMENT{Soft embedding}
    
    \STATE[] \textcolor{gray}{\# DLM's Autoregressive Pipeline:}
    %\STATE $\hat{\mathbf{y}} \leftarrow \tilde{f}_\theta(\mathbf{h})$ %\COMMENT{DLM's autoregressive pipeline}
    %\STATE $\hat{\Y} \leftarrow \tilde{f}_\theta(\mathbf p_\text{soft})$ \hfill \COMMENT{OR}
    %\STATE[9':] $\hat{\mathbf{y}} \leftarrow \tilde{f}_\theta(\mathbf p_\text{soft} \mid \tilde{\Y})$  \COMMENT{TF alternative:  $\tilde{\Y} = \text{one\_hot}(\Y)$}
    \STATE $\hat{\mathbf{y}} \leftarrow \tilde{f}_\theta(\mathbf p_\text{soft} \mid \tilde{\Y})$  \COMMENT{TF using  $\tilde{\Y} = \text{one\_hot}(\Y)$}
    %\addtocounter{ALG@line}{-1} % Don't increment the counter
    % \setcounter{ALG@line}{9}
    %\addtocounter{algocfline}{-1}
    \STATE[] \textcolor{gray}{\# Gradient-based updates}
    \STATE $\mathcal{L} \leftarrow d(\hat{\mathbf{y}}, \mathbf{y})$
    \STATE $\mathbf{Z}, \mathbf{\Phi} \leftarrow \text{Adam}(\mathbf{Z}, \mathbf{\Phi}, \nabla_{\mathbf{Z}} \mathcal{L}, \nabla_{\mathbf{\Phi}} \mathcal{L})$
\ENDFOR
\STATE \textbf{return} $\mathbf{Z}$
\end{algorithmic}
\end{algorithm}
}

\section{Method: Differentiable language models}
\label{sec:method}

Let $\X$ be a prompt containing a sequence of $N$ tokens and $\Y=f_\theta(\X)$ be the corresponding $M-$token output of a pretrained, frozen LM $f_\theta$ with parameters $\theta$. Denoting \V\ as the vocabulary over which $f_\theta$ is defined, then $\X\in\V^N$, $\Y\in\V^M$ and $f_\theta: \V^* \rightarrow \V^*$.

% Language Model Inversion (LMI) aims to recover $K$  $N$-token prompts $\X_i, \; i=1,\cdots,K$ given $K$ $M$-token target sequences $\Y_i$, such that
Given a target output \Y\ and a distance function $d: \V^* \times \V^* \rightarrow \mathbb{R}$ defined over the space of tokens, Language Model Inversion (LMI) aims to recover a prompt \X\ such that
\begin{align}
\,\, \X = \underset{\mathbf p\in \Voc^N}{\mathrm{argmin}} \; d( f_\theta(\mathbf p), \; \Y) 
\end{align}
% In this paper, we additionally define a difficulty term. \KS{I don't understand this. Todo: discuss with KD} \KD{Would it make more sense to frame the parameterised difficulty as an evaluation protocol device rather than a problem feature? We already have the greedy-decoding as an evaluation protocol device. It guarantees that we report a lower bound on the performance, and it removes the confounders related to choosing a stochastic decoding strategy (e.g. decoding temperature, top-k, top-p filters...). The addition of a difficulty parameterisation aims at giving insights about how the invertor will generalise to different kind of data (ID vs OOD), i.e. how it will perform when having to operate on data that are increasingly more out-of-distribution for the subject LM (because, while 'fluency' is defined using  perplexity, the perplexity of a given sentence is also a measure of similarity between said sentence and the pre-training distribution of the subject LM - lower perplexity means ID with respect to the pre-training distribution, and higher means OOD). }

\subsection{From tokens to distributions over tokens}
\label{sec:distributional-shift}

We redefine LMs from their traditional view as functions from sequences of prompt tokens \X\ to sequences of output tokens \Y. Instead, we treat it as a function $\ft_\theta: \Pd^*(\V) \rightarrow \Pd^*(\V)$ from and to the space of probability distributions $\Pd(\V)$ over vocabulary \V. We refer to $\ft_\theta$ as a \emph{Differentiable Language Model (DLM)}. In this context, rather than discussing input prompt \X\ or target sequence \Y, we deal with sequences of distributions over tokens (SDoT) $\Xt, \Yt \in\Pd^*(\V)$. 

The key insight is that it facilitates differentiation through the hard embedding and next-token sampling modules, which in turn leads to end-to-end differentiability. 
Given an SDoT input $\Xt$, we can swap the hard embedding module for a soft embedding module $\tilde{E}: \Pd(\Voc) \rightarrow \mathbb{R}^d$ which is implemented as:
$
\tilde{E}(\Xt_i) = \mathbf{W}_{E} \cdot \Xt_i
$
where $\mathbf{W}_E \in \mathbb{R}^{|\V| \times d}$ is the embedding matrix and the resulting implementation acts like a soft attention over the vocabulary embeddings, with weightings proportional to the likelihood of each token in the $i$-th distribution over tokens in the SDoT input $\Xt$.

% \KD{We need to define an input in the SDoT format before talking about the soft embedding as it is an input to the soft embedding function/module}
% \KS{What is the advantage of mentioning this here? Is it used anywhere?}
% \KD{It is used in the DLM pipeline inplace of the hard embedding module, I am updating Figure 1 middle to show that, sorry, I forgot about it being a placeholder}
% \KD{To update Figure 1 middle, I want to have the same notation as in the text, so I came back here to figure out how to notate it, but I am realising it is important to define the SDoT input, like $\tilde{x}\in\Pd(\Voc)$}

%\subsection{Sampling Tokens from DLMs}
\subsection{Gradients: Next-token samples, learnable prompts}
\label{subsec:sampling}
With the input interface of LMs addressed via the soft embedding module, it remains to update its output interface, i.e. the next-token sampling module. 
We use the Gumbel-Softmax (GS) reparameterization trick~\citep{jang2017categorical,maddison2017concrete} to build a GS-based next-token sampling module that is differentiable.
As detailed in Section~\ref{sec:background}, the GS distribution is designed to enable gradient-based optimisation through discrete, stochastic operations. 
%, it acts as a continuous relaxation of the \texttt{argmax} function by using a Gumbel-max trick combined with a temperature $\tau$ controlled softmax function. 
The GS distribution approaches the true categorical distribution it simulates as  $\tau \to 0$ or a uniform distribution as $\tau \to \infty$.

Moreover, within the context of the LMI problem, we represent the $N$-token long learnable prompt $\X\in\V^N$ as a sequence of categorical distributions over tokens with a sequence of unnormalized logits $\Z\in\mathbb{R}^{N\times V}$. 
We thus allow gradient backpropagation into $\Z$ by means of another GS gradient estimator.

\begin{comment} 
% \paragraph{Gumbel-Softmax Gradient Estimator with Learnable Temperature $\tau$.}
We use the Gumbel-Softmax (GS) distribution~\citep{jang2017categorical,maddison2017concrete}, which is a differentiable method to sample from a categorical distribution. Specifically designed to enable gradient-based optimisation through discrete, stochastic operations, it acts as a continuous relaxation of the \texttt{argmax} function by using a Gumbel-max trick combined with a temperature $\tau$ controlled softmax function. The distribution approaches a categorical distribution as  $\tau \to 0$ and a uniform distribution as $\tau \to \infty$. 

% This joint-parameterisation of the GS gradient estimator, with its Gumbel noise injection and learnable temperature parameter, bears resemblance to Langevin Monte Carlo methods, where noise is added to enable exploration of the probability landscape while temperature controls the trade-off between exploration and exploitation.

% Given logits $\mathbf{Z}\in\mathbb{R}^{N\times|\Voc|}$ and temperature $\tau > 0$, we obtain soft samples $\mathbf{p}_{\text{soft}}$ lying on the probability simplex.
% As $\tau \to 0$, the distribution approaches a categorical distribution, and as $\tau \to \infty$, it approaches a uniform distribution.
\end{comment}

\subsection{Learnable Temperature Parameters for Gradients}
% \subsection{ Gradient Estimation via Learnable \& Decoupled Temperature Parameters}
\label{sec:decoupled-learnable-temperatures}
Within the context of the GS gradient estimator over the learnable prompt $\X$ along, we follow ~\citet{Havrylov2017} and derive the temperature $\tau$ from a learnable parameter $\phi$, optimized alongside the learnable logits $\mathbf{Z}$ representing the learnable prompt $\X$, as 
$
    \tau_{\text{eff}} = \epsilon + \tau_0 \cdot (1 + \tanh(\phi))
$
where $\tau_0$ is an initial temperature parameter. 
This allows the optimization to automatically navigate the bias-variance trade-off inherent to GS-based gradient estimators. 
We expand this trick by decoupling the temperature for each learnable logit in $\mathbf{Z}$, thus learning a vector $\mathbf{\Phi}\in\mathbb{R}^N$, that enables automatically navigating the bias-variance trade-off for each logit's GS-based gradient estimation.

% \KS{Move to experiments
% For comparison, we also implement DLMI with the REINFORCE gradient estimator~\citep{williams1992simple} over loss function $\mathcal{L}$ as follows:
% \begin{equation}
%     \nabla_{\mathbf{Z}} \mathcal{L} \approx (\mathcal{L} - b) \cdot \nabla_{\mathbf{Z}} \log p_{\mathbf{Z}}(\mathbf{x})
% \end{equation}
% where $b$ is an exponential moving average baseline for variance reduction.
% While unbiased, REINFORCE is known to exhibit higher variance than GS-based estimators, but it is unclear how they relate in the context of DLMs.
% }

% \paragraph{Training vs.\ Inference Behavior.}
% The GS-based gradient estimation is only active during training when gradients must flow through the sampling operation.
% At inference time, DLMs behave identically to standard LMs: discrete tokens are sampled and no gradient computation is required.

% \KD{
% In 4.2 Sampling Tokens from DLMs, there is a possible issue: paragraph 1 talks about the GS distribution or reparameterization trick as it is used in a GS-based version of the Next-Token Sampling module (from Figure 1 middle - the bottom module), but then paragraph 2 talks about the GS distribution that is used over the parameterisation of the learnable prompts (detailing the decoupled learnable temperature). The problem is that this last concept is not linked to the DLM pipeline. It is linked to the 'gradient estimator' concept that is captured in Table 1's Grad. Est. column.
% }
% \KS{Add a comment at the end to clarify. Max 2 sentences.}

\subsection{Algorithm: Differentiable LM Inversion (DLMI)}
Our approach combines four elements: The distributional viewpoint via soft embedding \& GS-based next-token sampling; GS gradient estimator for the learnable prompt;  decoupled, learnable temperature parameters; and and Teacher Forcing (TF - ~\citet{williams1989TeacherForcing-learning}).

We jointly optimise learnable input logits $\mathbf{Z} \in \mathbb{R}^{N \times |\Voc|}$ and learnable parameters $\mathbf{\Phi}\in\mathbb{R}^N$ such that a frozen LM $f_\theta$ produces a given target output $\Y\in\Voc^*$, via gradient descent through its differentiable counterpart $\ft_\theta$. 
The input logits represent the $N$-token prompt to be recovered and $\mathbf{\Phi}$ (one for each input logit's GS gradient estimation) is used to derive decoupled temperatures $\mathbf{\tau}\in\mathbb{R}^N$.

Algorithm~\ref{alg:dlmi} presents the optimisation procedure with a GS gradient estimator on the sampling computational branch that stems from the learnable logits $\Z$. 
% 
% \KD{
% This last sentence is not correct: in the DLMI algorithm, with TF, the GS-based next-token sampling is essentially skirted, because its two main features are neutralised, towit no Gumbel noise injection and $\tau=1$ means no differences with respect to the distribution that came out of the Vocabulary Projection module from figure 1.
% }

% \KD{
% Rather, in algo 1's title, the replace-able part that ought to be highlighted is the GS gradient estimator on the sampling computational branch that stems from the unnormalized logits $\Z$ (as opposed to e.g. REINFORCE gradient estimator on that same branch - which is what the REINFORCE-labelled method in previous version of the figures 2 and 3 was implementing).
% }

% \KD{
% I think the sentence should rather say:
% 'we present the optimisation procedure with a GS gradient estimator on the sampling computational branch that stems from the unnormalized logits $\Z$'
% }
% 
Line $10$ implements Teacher-Forcing (TF) where the returned distribution over tokens $\hat{\Y} \in\Pd^M(\V)$ uses a $\tau=1$-tempered softmax operation without Gumbel noise injection. 
In the ablation study, we test this algorithm without TF (line 10 is replaced with $\hat{\Y}$ computed as the softmax over the unnormalized logits return by the Vocabulary Projection module - cf. Figure~\ref{fig:DLMI}).

{\scriptsize
\begin{algorithm}[h]%[thbp]
\caption{DifficultyDrivenTargetGeneration}
\label{algo:target-generation}
\begin{algorithmic}[1]
\REQUIRE LM $f_\theta$ defined over vocabulary $\V$, difficulty $k$, length $M$, noise $\sigma>0$
\ENSURE Target sequence $y$ of length $M$
\vspace{1em}
\STATE $\Y_0 \leftarrow \{\text{BoS token}\}$
\FOR{$i = 1$ to $M$}
    \STATE $A \leftarrow \{ \Y_j\}_{j=0}^{i-1}$ \hfill \COMMENT{all tokens thus far}
    \STATE $\mathbf{p} \leftarrow \delta f_\theta(A)$ \hfill \COMMENT{yields distr. over tokens}
    \STATE Sort tokens by probability $\mathbf{p}_i$: $[t_1, t_2, \ldots, t_{|V|}]$
    %\STATE $\Y_i \leftarrow$ SampleTokenWindow($[t_{k-w}, t_{k+w}]$)    
    %\STATE $|V'| \leftarrow |\{v : \mathbf{p}_v > 0\}|$ \hfill \COMMENT{Effective vocabulary size}
    \STATE Sample $k' \sim \mathcal{N}(k, \sigma^2)$ and round to nearest integer
    \STATE $k' \leftarrow \text{clip}(k', 1, |\V|)$ \hfill \COMMENT{Ensure $k' \in [1, |\V|]$}
    \STATE $\mathbf{y}_i \leftarrow t_{k'}$ \hfill \COMMENT{Select token at perturbed rank $k'$}
    \IF{$\mathbf{y}_i = \text{EoS}$}
        \STATE Pad \Y\ until length $M$ with EoS token
        \STATE \textbf{break} \hfill \COMMENT{Terminate on end-of-sequence token}
    \ENDIF
    %\STATE $i \leftarrow i + 1$
\ENDFOR
\STATE \textbf{return} $\Y = \{\Y_i\}_{i=1}^M$
\end{algorithmic}
\end{algorithm}
}

\section{Experiments}
\label{sec:invertibility-evaluation}
\label{sec:experiments}

% 2 x 2 version
\begin{figure*}[t]
    \centering
    \begin{tabular}{@{}c@{}c@{}}
         $256$ Opt. Steps & $2048$ Opt. Steps \\ 
         \includegraphics[width=0.49\linewidth]{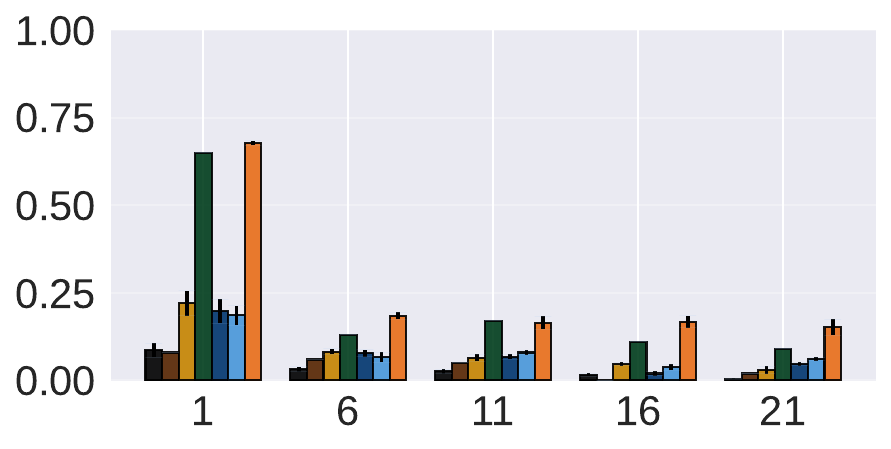} & \includegraphics[width=0.49\linewidth]{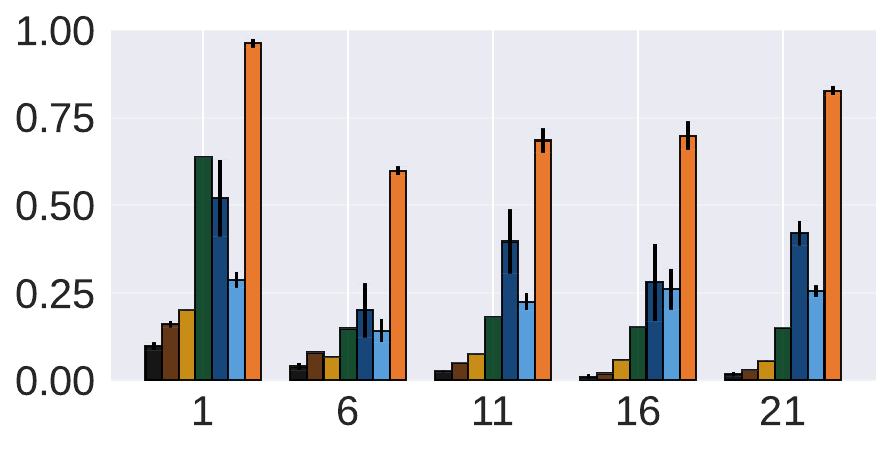} \\ %\hline 
         \includegraphics[width=0.49\linewidth]{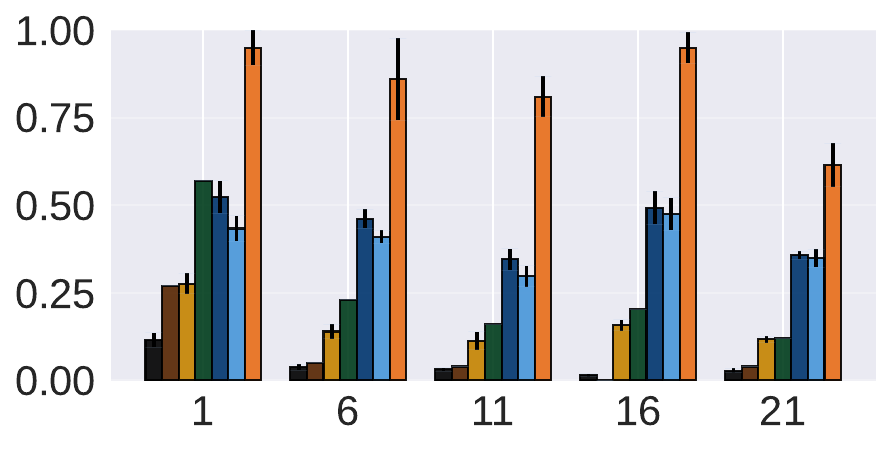} & \includegraphics[width=0.49\linewidth]{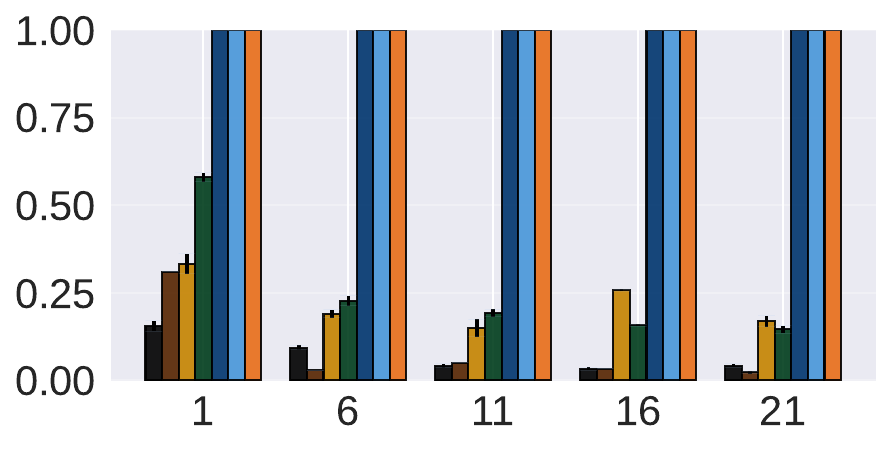} \\ %\hline 
         %\multicolumn{2}{c}{\includegraphics[width=.65\linewidth]{figures/LCSRatios_avg_lcs_ratio_legend-2lines.pdf}} \\ %\hline
         \multicolumn{2}{c}{\includegraphics[width=\linewidth]{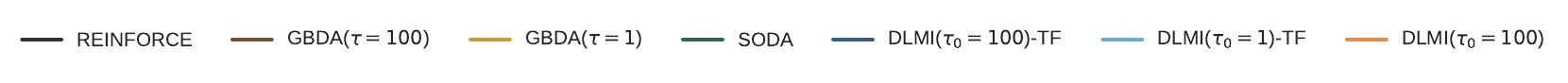}} \\ %\hline
    \end{tabular}
    \caption{Plots of accuracy (LCS Ratios) vs target difficulty $k$ for SmolLM2-135M (top) and SmolLM3-3B (bottom) results after 256 (left) or 2048 (right) optimization steps with $N=80$ and $M=20$.%\KS{The legend should be fit in one row to save vertical space.}
    }
    \label{fig:placeholder-results-pl80}
\end{figure*}

\subsection{Dataset for evaluation}
\label{sec:dataset-generation}
\begin{comment}
We adopted two methods for generating evaluation data. The straightforward approach was to generate random ground truth prompt data $\bar{\X}$, obtain the target via $\Y = f_\theta(\bar{\X})$ then apply Algorithm DLMI to recover an optimised input prompt \X. We used this in \KS{reference tables and figures}. 
\KD{
If by straightforward approach we mean to refer to the previous works' evaluation protocol (as detailed in related works and background), then the target is obtained via $\Y = f_\theta^{\mathrm{argmax}}(\bar{\X})$, which is a notation I mean to propose to specify that the next-token sampling module is argmax, when using a common LM.
}
\end{comment}

To evaluate the performance of DLMI across varied difficulties, rather than to begin with ground-truth prompts like in previous works, we use an integer $k$ to control the generation of target outputs. 
Setting $k=1$ generates high probability (and therefore low perplexity, or low difficulty) target outputs while larger values of $k$ imply higher difficulty. 
We implemented this by starting from a beginning-of-sentence (BoS) token and autoregressing one token at a time. 
%That is, in each iteration, we sort the tokens in \V\ based on the probabilities yielded by $\ft_\theta$ and sample a token from a narrow window around the token with the $k^{th}$ largest probability. 
% \KD{
% $\ft_\theta$ can indeed be used here, but it would create bias into the generated outputs due to the Gumbel noise injection. In practice, what is used in not defined in the paper. So I think that maybe we need to define a partial, single-pass (not loop) LM where the output is the distribution outputted by the Vocabulary Projector module in the form of unnormalized logits. e.g. $\delta f_\theta : \V^* \rightarrow \Pd^(\V)$ which contains only one call to the embedding module + positional encoding + transformer module + vocabulary projection module, leaving the next-token sampling module out.
% }
However, this single autoregressing step relies on $\delta f_\theta : \V^* \rightarrow \Pd^(\V)$ which is the subpart of the autoregressive LM pipeline described in Figure~\ref{fig:DLMI}, that contains only one call to the embedding module, the positional encoding, the transformer module, and the vocabulary projection module (leaving the next-token sampling module out), plus a softmax operation to transform the vocabulary projection's unnormalized logits into a distribution over tokens.
Thus, in each iteration, we sort the tokens in \V\ based on the probabilities yielded by $\delta f_\theta$ and sample a token from a narrow window around the token with the $k^{th}$ largest probability. 
This token is added to the cumulative output ($A$ in Algorithm~\ref{algo:target-generation}) and the process is repeated $M$ times to obtain a target output \Y\ of length $M$. 
This is summarised in Algorithm~\ref{algo:target-generation}.
%\KS{Explain the sampling method used by line 6 of Algorithm.}

For each subject LM that we study in Section~\ref{sec:experiments}, we evaluate over a generated dataset of target outputs of size $25$ spread over difficulties $k\in[1,6,11,16,21]$, with $5$ target output samples for each value of $k$.

\begin{comment}
% 1 x 4 version
\begin{figure*}[t]
    \centering
    \begin{tabular}{c|c|c|c}
        \multicolumn{2}{c}{SmolLM2-135M} & \multicolumn{2}{c}{SmolLM3-3B} \\ \hline
         $256$ Opt. Steps & $2048$ Opt. Steps  & $256$ Opt. Steps & $2048$ Opt. Steps \\ \hline
         \includegraphics[width=0.24\linewidth]{figures/s2-pl80-LCSRatios_avg_lcs_ratio_conditioned_barplot.png} & \includegraphics[width=0.24\linewidth]{figures/s2-pl80-LCSRatios_avg_lcs_ratio_conditioned_barplot.png} & 
         \includegraphics[width=0.24\linewidth]{figures/s2-pl80-LCSRatios_avg_lcs_ratio_conditioned_barplot.png} & \includegraphics[width=0.24\linewidth]{figures/s2-pl80-LCSRatios_avg_lcs_ratio_conditioned_barplot.png} \\ \hline 
         \multicolumn{4}{c}{Legend Placeholder} \\ \hline
    \end{tabular}
    \caption{SmolLM2-135M (top) and SmolLM3-3B (bottom) results after 256 (left) or 2048 (right) optimization steps with $N=80$.}
    \label{fig:placeholder-results-pl80}
\end{figure*}
\end{comment}

\begin{figure*}[t]
    \centering
    \begin{tabular}{@{}c@{}c@{}}
         $256$ Opt. Steps & $2048$ Opt. Steps \\ 
         \includegraphics[width=0.49\linewidth]{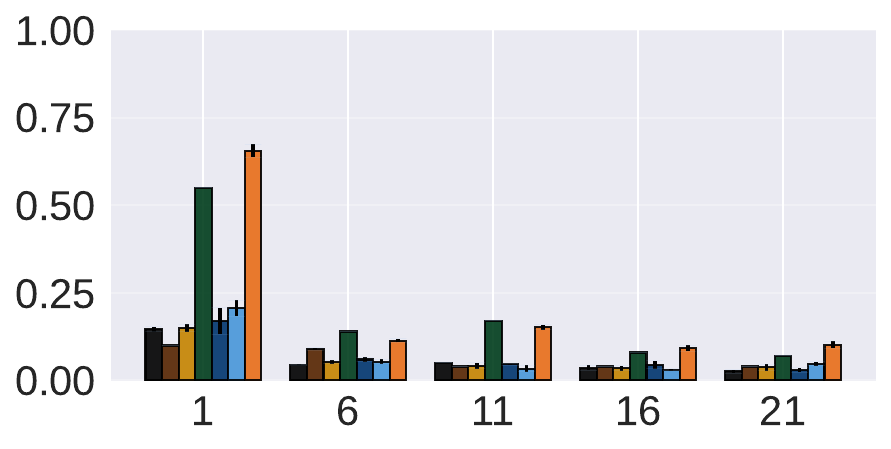} & \includegraphics[width=0.49\linewidth]{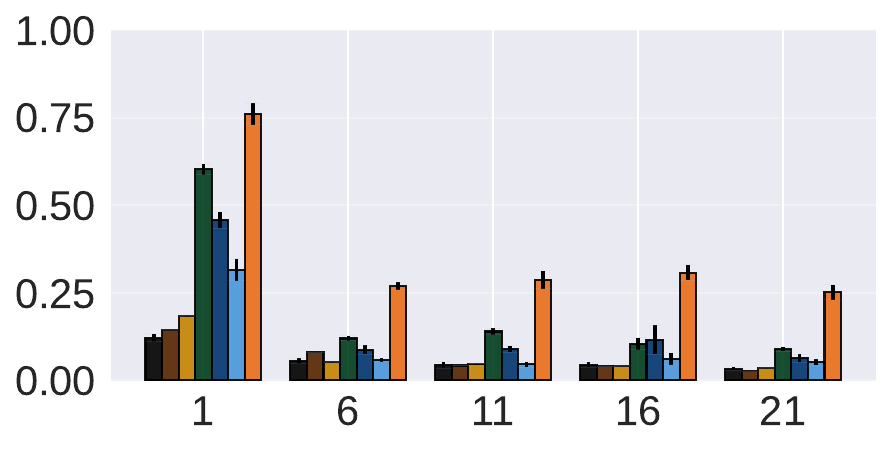} \\ %\hline 
         \includegraphics[width=0.49\linewidth]{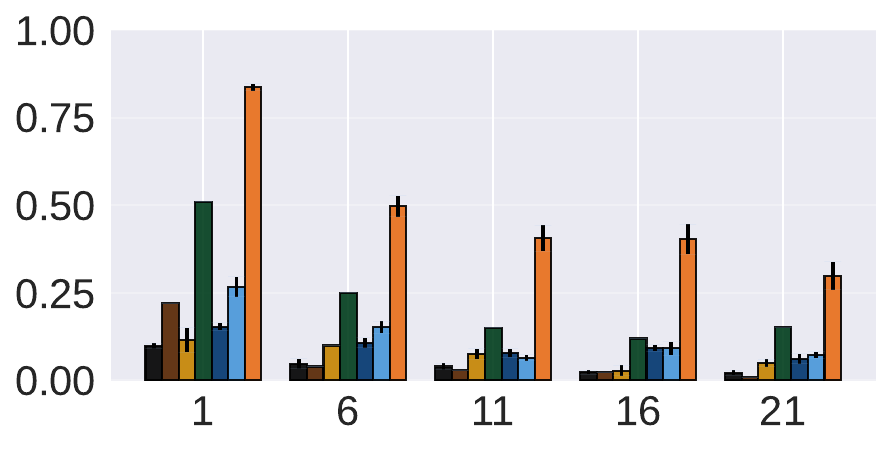} & \includegraphics[width=0.49\linewidth]{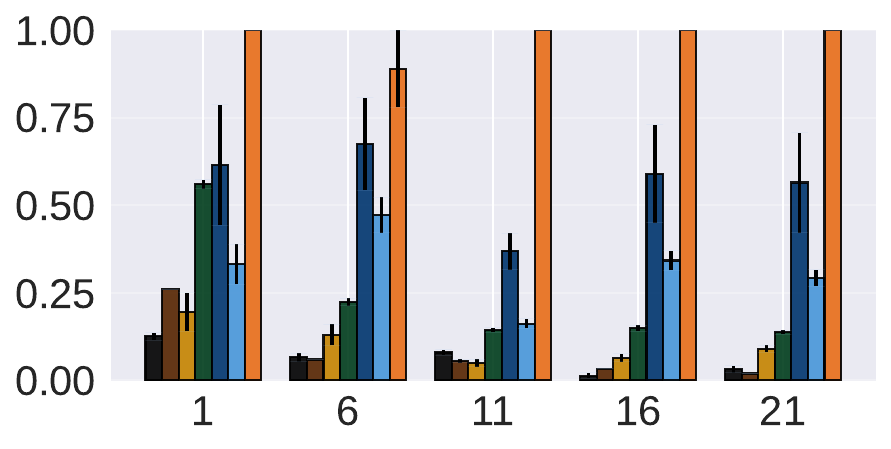} \\ %\hline 
         %\multicolumn{2}{c}{\includegraphics[width=.65\linewidth]{figures/LCSRatios_avg_lcs_ratio_legend.pdf}} \\ %\hline
         \multicolumn{2}{c}{\includegraphics[width=\linewidth]{figures/LCSRatios_avg_lcs_ratio_legend.pdf}} \\ %\hline
    \end{tabular}
    \caption{Plots of accuracy (LCS Ratios) vs target difficulty $k$ for SmolLM2-135M (top) and SmolLM3-3B (bottom) results after 256 (left) or 2048 (right) optimization steps with $N=10$ and $M=20$. %\KS{The legend should be fit in one row to save vertical space.}
    }
    \label{fig:placeholder-results-pl10}
\end{figure*}

\begin{comment}
% 1 x 4 version
\begin{figure*}[t]
    \centering
    \begin{tabular}{c|c|c|c}
        \multicolumn{2}{c}{SmolLM2-135M} & \multicolumn{2}{c}{SmolLM3-3B} \\ \hline
         $256$ Opt. Steps & $2048$ Opt. Steps  & $256$ Opt. Steps & $2048$ Opt. Steps \\ \hline
         \includegraphics[width=0.24\linewidth]{figures/s2-pl80-LCSRatios_avg_lcs_ratio_conditioned_barplot.png} & \includegraphics[width=0.24\linewidth]{figures/s2-pl80-LCSRatios_avg_lcs_ratio_conditioned_barplot.png} & 
         \includegraphics[width=0.24\linewidth]{figures/s2-pl80-LCSRatios_avg_lcs_ratio_conditioned_barplot.png} & \includegraphics[width=0.24\linewidth]{figures/s2-pl80-LCSRatios_avg_lcs_ratio_conditioned_barplot.png} \\ \hline 
         \multicolumn{4}{c}{Legend Placeholder} \\ \hline
    \end{tabular}
    \caption{SmolLM2-135M (top) and SmolLM3-3B (bottom) results after 256 (left) or 2048 (right) optimization steps with $N=10$.}
    \label{fig:placeholder-results-pl10}
\end{figure*}
\end{comment}
%\subsection{Evaluation Setup \& Results}
\subsection{Comparisons}

\paragraph{REINFORCE Gradient Estimator.} 
For comparison, we also implement DLMI with the REINFORCE gradient estimator~\citep{williams1992simple} for the learnable logits \Z, over loss function $\mathcal{L}$, as follows:
\begin{equation}
    \nabla_{\mathbf{Z}} \mathcal{L} \approx (\mathcal{L} - b) \cdot \nabla_{\mathbf{Z}} \log p_{\mathbf{Z}}(\mathbf{x})
\end{equation}
where $b$ is an exponential moving average baseline for variance reduction.
While unbiased, REINFORCE is known to exhibit higher variance than GS-based estimators, but it is unclear how they relate in the context of DLMs.

\newcommand{\cmark}{{\color{green}\ding{51}}}%
\newcommand{\xmark}{{\color{red}\ding{55}}}%
% There was another definition of these below, I have removed them now..

\begin{table}[htbp]
\centering
\small
\begin{tabular}{l|ccccc}
\toprule
Methods 
& Soft Emb. 
& Grad. Est. 
& DL-$\tau$ 
& TF \\
\midrule
DLMI($\tau_0=X$) 
& \cmark & GS & \cmark & \cmark \\
\midrule 
DLMI($\tau_0$$=$X)-TF & \cmark & GS & \cmark & \xmark \\

REINFORCE 
& \xmark & REINFORCE & N/A & \xmark \\

%DLMI($\tau=X$) & \cmark & GS & \cmark & \xmark \\

%GS($\tau=X$) (GBDA) & \cmark & GS & \xmark & \xmark \\
GBDA($\tau=X$) & \cmark & GS & \xmark & \xmark \\

SODA ($\tau=X$) & \cmark & \xmark & \xmark & \cmark \\
\bottomrule
\end{tabular}
\caption{Comparing DLMI variants and previous works.}
\label{tab:dlmi_ablations}
\end{table}

\paragraph{GBDA.}
Because our approach shares many building blocks with the \textbf{GBDA} approach from ~\citet{guo-etal-2021-GBDA-EMNLP2021}, we include our LMI-adapted re-implementation of it for comparison. 
As detailed in Table~\ref{tab:dlmi_ablations}, \textbf{GBDA} acts as an ablation of our DLMI approach, removing the decoupled and learnable temperature parameters $\Phi$ and the TF trick.

\paragraph{SODA.}
We also include the state-of-the-art \textbf{SODA} inverter from ~\citet{skapars2025SODA-gpt}. 
Both SODA and DLMI optimize learnable logits \Z\ that represents the prompt \X\ via gradient descent, but SODA only relaxes the computational graph with the soft embedding module and does not 'simulate' the sampling of a prompt from the categorical distributions defined by the learnable logits (i.e. it does not rely on a gradient estimator). 
While the authors proposed to use it with $10000$ optimisation steps, we observed no improvement of the optimised prompt \X\ beyond ~200 steps, and therefore present results in a fair comparison with other methods after different number of optimisations steps.

\subsection{Evaluation Setup \& Results}

\paragraph{Metrics.}
We measure the accuracy at finding a prompt \X\ for a target output \Y, given the subject LM, using token-level longest-common-subsequence (LCS) ratios (similar to the ROUGE-L metric~\citep{lin2004rouge}), and we report how this accuracy varies as the difficulty of the target outputs changes (i.e. as the average rank $k$ over generated targets output increases). 
%Our measure of invertibility of a given DLM under our GS-based gradient estimators is the area-under-the-curve of average LCS vs $k$. 

Crucially, we measure the LCS ratio between \textit{greedily-decoded} outputs $\bar{\Y} = f_\theta^{\mathrm{argmax}}(\X)$ and target outputs \Y, in order to guard our results from confounders in the next-token sampling module, of which many flavours exist. 
This choice is motivated by the aim to provide a deterministic measure of the lower-bound estimation of the performance of the evaluated inverters. 

% \todo[inline]{Discuss the cheating strategy consisting of an inverse prompt containing the target output in the form of an instruction: e.g. ``Repeat the following sentence: `TARGET\_OUTPUT' ''.
% }
We note that a trivial ``cheating'' strategy exists whereby an optimised prompt \X\ could contain the target output \Y\ in the form of an instruction, e.g., ``Repeat the following sentence: `TARGET\_OUTPUT'\,''. 
We verify a-posterio the token overlap rate between \X\ and \Y\ for all methods and observed rates well below $20\%$, guaranteeing that none of the method converged onto this ``cheating'' strategy.
%\KD{maybe we can remove this paragraph? or I will try to add a barplot to the appendix to illustrate}

%OPTIONAL: Discuss baseline Textgrad performance.

% \item{
% OPTIONAL: different generation scheme: argmax (default) vs sampling (given temperature, top\_p, top\_k ... --> requires setting up a threshold to frequency of target output being obtained ) vs beam search (e.g. CoT-promoting scheme --> also requires setting up a threshold as it involves sampling)
% }
% We also report a measure of efficiency by measuring the average number of optimisation steps required to achieved the reported accuracies, out of a budget of $2000$ steps, with std. errors: e.g. AuC of $70\pm5\%$ @ $1800\pm50$ optimisation steps. Or on an 2D graph with x-axis being the average number of optimisation steps and the y-axis being the AuC percentage, with the points being represented by ellipses whose diameters are parameterised by the std.errors. 
% REMOVED: because it does not bring anything compared to SmolLM2 at comparable size
% \item{
% OPTIONAL: Results with DistilGPT2 (88M parameters).
% }
\paragraph{Language Models \& Hardware.}
We experiment with LMs at different parameter scales, with SmolLM2-135M and SmolLM3-3B-Base\footnote{SmolLM2/3 family:\url{https://huggingface.co/collections/HuggingFaceTB/smollm2} \& \url{https://huggingface.co/HuggingFaceTB/SmolLM3-3B-Base}}. 
We run our experiments on separate NVIDIA RTX A6000 48.0 GB GPUs, NVIDIA A100 40GB GPUs, and AMD Radeon 8060S graphics. 
Unless locally specified, we run $5$ seeds for each method. 
Hyperparameters can be found in Table~\ref{tab:hyperparams}.

%\item{
% Results with Qwen3 family of models\footnote{Qwen 3 family: \url{https://huggingface.co/collections/Qwen/qwen3}}: 0.6B vs 1.7B vs 4B vs 8B vs 14B vs 30/32B (base vs instruct when available)
% }
\begin{comment}
\paragraph{LMI Performance.}
Figures~\ref{fig:placeholder-results-pl80} and ~\ref{fig:placeholder-results-pl10} show the measured LCS ratios as a function of the target output difficulty $k\in[1,6,11,16,21]$ for, respectively, $N=80$ and $N=10$. 

Results as the prompt lengths scales ($PL=80$ default vs 10) 

Results as the LM parameter size scales

\paragraph{Gradient Estimators' Bias \& Variance.}
Figure~\ref{fig:placeholder-gradient-var-bias}(top) shows mean$\pm$std. err. over random seeds and samples

Discussion regarding the variance and bias of the gradient estimators used, in comparison to the REINFORCE gradient estimator ((unbiased but high-variance - \citet{williams1992simple}).

How do the different gradient estimators behave in terms of variance and bias?    

How does the DLM paradigm perform in the context of the LMI problem?
\end{comment}

% 2 x 2 version
\begin{figure}[t]
    \centering
    \begin{tabular}{c}
         \includegraphics[width=\linewidth]{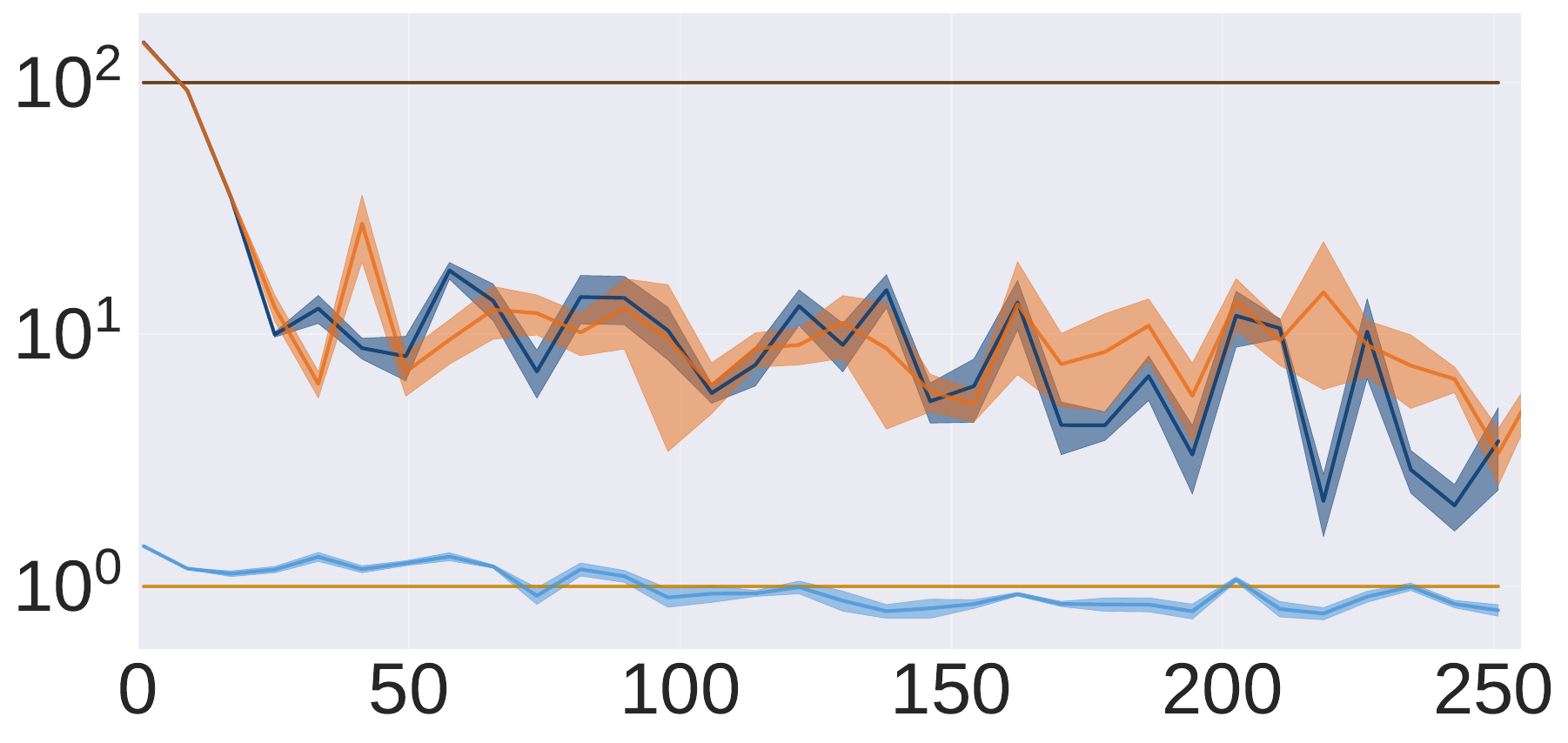}  \\ %\hline 
         \includegraphics[width=\linewidth]{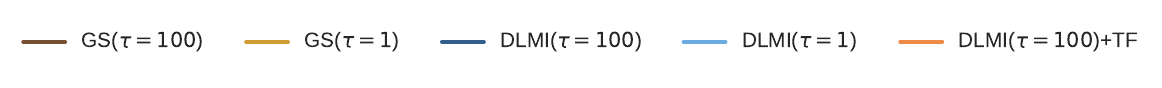}
         %\multicolumn{2}{c}{Legend Placeholder} \\ \hline
    \end{tabular}
    \caption{ Mean $\pm$ std.err. of the (mean over prompt token - when decoupled) effective temperature $\tau$ for SmolLM2-135M over $256$ optimization steps, with $N=80$.}
    \label{fig:placeholder-temperature}
    \label{fig:placeholder-gradient-var-bias}
\end{figure}

% 2 x 2 version
\begin{figure*}[t]
    \centering
    \begin{tabular}{@{}c@{}c@{}}
         Max Grad. Var. & Max Grad. Bias \\ 
         \includegraphics[width=0.49\linewidth]{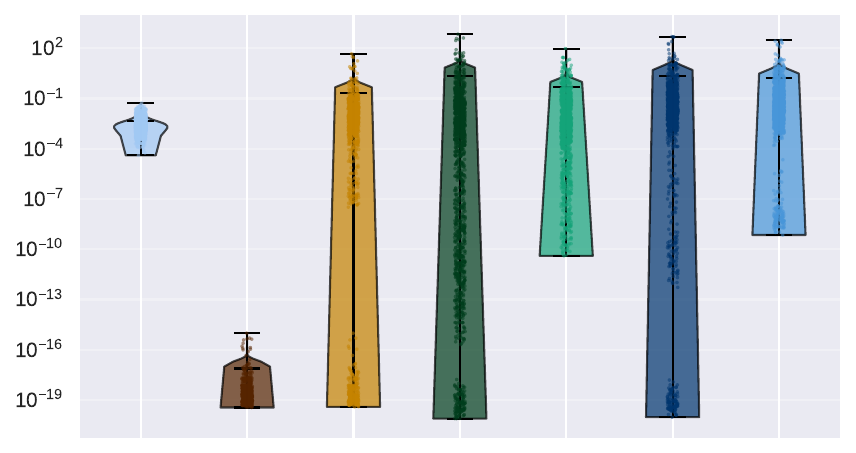} & \includegraphics[width=0.49\linewidth]{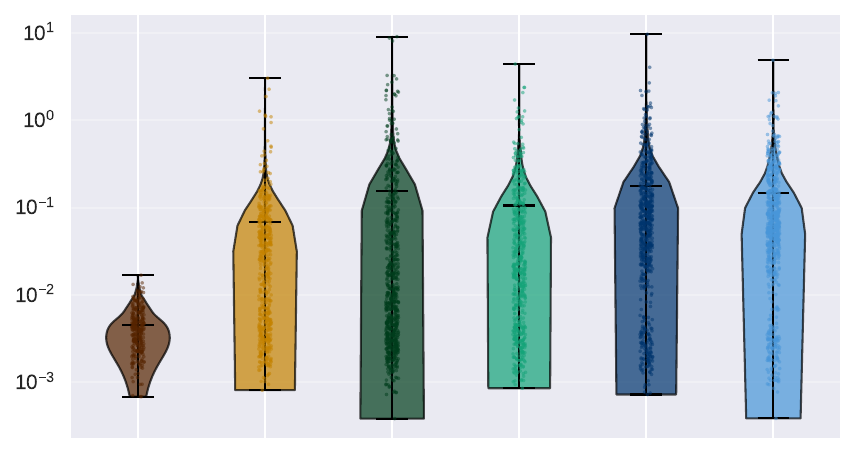} \\ %\hline 
         %\includegraphics[width=0.425\linewidth]{figures/s3-pl80-256-LCSRatios_avg_lcs_ratio_conditioned_barplot.pdf} & \includegraphics[width=0.425\linewidth]{figures/s3-pl80-2048-LCSRatios_avg_lcs_ratio_conditioned_barplot.pdf} \\ %\hline 
         %\multicolumn{2}{c}{\includegraphics[width=.65\linewidth]{figures/LCSRatios_avg_lcs_ratio_legend-2lines.pdf}} \\ %\hline
         \multicolumn{2}{c}{\includegraphics[width=\linewidth]{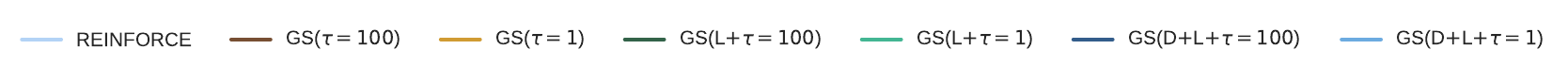}} \\ %\hline
    \end{tabular}
    \caption{Violin plots (with mean and extrema) in log scale of maximal gradient variance (left) and maximal gradient bias (right) for SmolLM2-135M over 256 optimization steps with $N=80$ and $M=20$ (maximal is taken over learnable logits $M\cdot |\V|$ at each optimization step). \textbf{D} stands for decoupled learnable temperature, \textbf{L} stands for learnable temperature. %\KS{The legend should be fit in one row to save vertical space.}
    }
    \label{fig:gradient-var-bias-results-pl80}
\end{figure*}
% Draft for Sections 4.3 and 5 of the DLMI Paper
% Language Model Inversion through End-to-End Differentiation

\textbf{LMI Performances.} Figures~2 and 3 show the measured LCS ratios as a function of the target output difficulty $k \in [1, 6, 11, 16, 21]$ for, respectively, $N = 80$ and $N = 10$, after both $256$ (left) and $2048$ (right) optimisation steps. 
Our proposed method achieves state-of-the-art results in all settings, and even near-perfect or perfect results, depending on the LM parameter size, when provided with longer prompt length ($N=80$) and bigger optimisation budget ($2048$ steps). 
As expected, our ablation called REINFORCE, which replaces the GS gradient estimator with the REINFORCE gradient estimator, yields very poor results. This shows that using GS-based gradient estimator is critical. However, it is not sufficient to use the GS gradient estimator in its fixed-temperature form, as the GBDA baseline (which also acts as an ablation where decoupled learnable temperature is removed) performs barely better than REINFORCE and well-below the performance levels of DLMI-TF, especially as the difficulty level $k$ increases. This shows that decoupled learnable temperature is the most critical element to enable practically-efficient LM inversion via DLMs. 
Comparing DLMI-TF (i.e. w/o TF) to DLMI then shows how the teacher-forcing trick increases sample-efficiency.

\textit{Results as the prompt length scales ($N = 80$ vs $N = 10$).} Comparing Figures~\ref{fig:placeholder-results-pl80} and ~\ref{fig:placeholder-results-pl10} reveals that prompt length substantially affects inversion performance. 
With longer prompts ($N = 80$), our DLMI approach demonstrates strikingly superior performance, particularly at higher difficulty levels where other methods can barely improve from below $~20\%$. 
The extended prompt length provides greater optimisation flexibility, allowing the learnable logits to explore a richer space of potential solutions. 
In contrast, shorter prompts ($N = 10$) constrain the optimisation landscape, leading to reduced performance across all methods. 
%Notably, the performance gap between DLMI and others (GBDA, SODA) widens as prompt length increases, suggesting that our method more effectively exploits additional optimisation capacity. 

%At $k = 1$ (low difficulty) with $N = 80$ and $2048$ optimisation steps, DLMI($\tau_0 = 100$) achieves perfect inversion (LCS $\approx 0.95$), whereas at $k = 21$ (high difficulty), performance degrades to approximately 0.75---still substantially above the baselines which struggle to exceed 0.50.

\textit{Results as the LM parameter size scales.} Our experiments across SmolLM2-135M and SmolLM3-3B reveal an interesting pattern: contrary to expectations, larger models are easier to invert. 
This observation is slightly orthogonal to findings from \citet{skapars2025SODA-gpt}, who report that vocabulary size and input length have greater impact on inversion difficulty than model parameter count alone. Critically, our DLMI approach maintains robust performance across both scales, as well as across difficulty levels (on the contrary to SODA), demonstrating the generalisability/external validity of the DLM paradigm.

\textbf{Gradient Estimators' Bias \& Variance.} Figure~\ref{fig:gradient-var-bias-results-pl80} shows violin plots (with mean and extrema) over random seeds and samples for the gradient variance and bias of the evaluated estimators.

%The REINFORCE estimator, while theoretically unbiased \citep{williams1992simple}, exhibits substantially higher variance compared to the Gumbel-Softmax (GS) based approaches, confirming the classical bias-variance trade-off documented in the literature \citep{bengio2013estimating}. Among GS-based methods, we observe that fixed-temperature variants (GS($\tau = 1$) and GS($\tau = 100$)) display characteristic behaviours: lower temperatures ($\tau = 1$) produce samples closer to the true categorical distribution but with higher gradient variance, while higher temperatures ($\tau = 100$) yield smoother but more biased gradients.

%Our DLMI variants with learnable temperatures demonstrate a crucial advantage: the decoupled, per-token temperature parameters $\Phi$ automatically navigate this bias-variance trade-off during optimisation. Figure~4(bottom) illustrates the evolution of the effective temperature $\tau_{\text{eff}}$ over optimisation steps, revealing that the learned temperatures adapt dynamically---typically starting high for exploration and gradually decreasing to refine the solution. This adaptive behaviour explains DLMI's superior performance: it combines the low-variance benefits of high-temperature GS sampling during early optimisation with the low-bias precision of low-temperature sampling as convergence approaches.

\textit{How do the different gradient estimators behave in terms of variance and bias?} The empirical results confirm theoretical predictions regarding the REINFORCE gradient estimators, as it provides statistically-significantly higher variance gradients than GS-based estimators, resulting in a much slower optimisation process which struggle to converge (as seen in Figures~\ref{fig:placeholder-results-pl80} and ~\ref{fig:placeholder-results-pl10}). 
However, fixed-temperature GS estimator does not exhibit the expected trade-off---lower $\tau$ reduces bias at the cost of increased variance---, but rather increasing $\tau$ ($\tau=100$) shows both statistically-significantly lower variance and bias than at lower value of $\tau$ ($\tau=1$). 
Then, adding learnable (\textbf{L}) and decoupled (\textbf{D}) temperature features yield variances and biases whose ranges are on par with that at the lower value of fixed-temperature (\textit{GS($\tau=1$)}). 
However, while their biases are statistically-indistinguishable from one value of $\tau_0$ to another, the range of their variances is smaller when $\tau_0=1$ compared to when $\tau_0=100$, and marginally on the higher end. 
This suggests that, in the context of DLMs, the GS-estimated gradients exhibit, at the very least, better variance properties when the temperature is kept higher throughout optimization (fixed temperature $\tau=100$), when it is starting at a higher $\tau_0=100$ and then learned jointly through the optimization.
Those results would suggests that a higher $\tau$ or $\tau_0$ would be beneficial in the context of DLMs, and our convergence results indeed correlate with this observations, as seen in Figure~\ref{fig:placeholder-results-pl80} where our method \textit{DLMI($\tau_0=100$)-TF} performs marginally better than \textit{DLMI($\tau_0=1$)-TF}.

\textit{How does the DLM paradigm perform in the context of the LMI problem?} The DLM paradigm---viewing LMs as functions over sequences of distributions over tokens (SDoTs) rather than sequences of tokens (SoTs)---proves highly effective for LMI. By enabling end-to-end differentiability, our approach unlocks end-to-end, flexible gradient-based optimisation where previous methods relied on discrete search (SODA, GCG) or fixed continuous relaxations (GBDA). The combination of (i) soft embeddings, (ii) GS gradient estimation with (iii) decoupled learnable temperatures, and (iv) Teacher Forcing yields consistent improvements across all difficulty levels and model scales. The anytime nature of our algorithm further enhances its practical utility: even with limited computational budgets (e.g., 256 optimisation steps), DLMI achieves competitive LCS ratios, with performance improving to new state-of-the-arts as additional steps are allocated.

\subsection{Additional results}

\paragraph{Qualitative example} 
\begin{comment} 
To provide a qualitative perspective, given a target output $\Y=\text{``ICML 2026 is expected to have a great array of papers''}$, our DLMI algorithm optimised an $N$$=$$80$-token-long prompt \X\  on SmolLM3-3B-Base to arrive at the prompt: 
\begin{figure}[htbp]
\includegraphics[width=\linewidth]{figures/text2.png}
\end{figure}
%
% ``MongoDB Judges언戸�Enumerable脚 LISTSlice المج� � hometown LOAD�� sv档 由 miềnwis Goλογ済渐อท fortune cors 슬 ditch الج Lewтор Moff Morrison coronęż delightful Sloven residential_BEGINmadan resumesIMPLIED OSP qualifier Cornell.device strand나는ischen响 spider beginnings危惠 EF discount규 spraw 식 contingent_span Gong BEEN genius guild impaired egregious Wireδιο碼 suspected Communic preferably DISPLAY playwright hereby播lst LSM''. 

The optimisation took $185$ steps, and was performed under 18mins.
\end{comment}
We show in Tables~\ref{tab:optimization_results_s2} and ~\ref{tab:optimization_results_s3} qualitative results of the learned prompts \X\ and resulting \textit{greedily-decoded} outputs $\bar{\Y} = f_\theta^{\mathrm{argmax}}(\X)$ given target outputs \Y after $2048$ optimization steps for, respectively SmolLM2-135M and SmolLM3-3B-base, with $N=10$ and $M=20$ (corresponding to Figure~\ref{fig:placeholder-results-pl10}(right)).

\paragraph{From LMI to Prompt Engineering.}
The LMI problem can be extended to the prompt engineering problem by considering only a partially-learnable prompts \X\ that contains some fixed and not-learnable conditioning tokens, for instance representing the question in a math world problem. 
We ran preliminary experiments in the context of prompt engineering where the learnable-part of the prompt is optimised with our DLM-powered gradient descent algorithm. 
Our results on the GSM8k benchmark~\citep{cobbe2021training-gsm8k} show promising improvements between $9-35\%$ relative increase, depending on the LM parameter size and family.

\section{Conclusion} 

We have demonstrated that language models can be made end-to-end differentiable through a shift in perspective---treating them as functions over distributions rather than discrete tokens---combined with Gumbel-Softmax gradient estimation and learnable temperature parameters. The resulting DLMI algorithm achieves state-of-the-art performance on the language model inversion problem across a fluency-parameterised evaluation protocol. More broadly, this work establishes DLMs as a foundation for gradient-based prompt optimisation, with applications spanning adversarial auditing, controllable generation, and interpretability research.

\section*{Acknowledgment}
This work was funded by ELIAI (The Edinburgh Laboratory for Integrated Artificial Intelligence) EPSRC (grant EP/W002876/1). 
We thank Sean Memery for helpful discussions in the early stage of the project.

\section*{Impact Statement}
This paper presents work whose goal is to advance the field of Machine Learning. 
There are many potential societal consequences of our work, none which we feel must be specifically highlighted here.

% \todo[inline]{
% Authors are \textbf{required} to include a statement of the potential broader
% impact of their work, including its ethical aspects and future societal
% consequences. This statement should be in an unnumbered section at the end of
% the paper (co-located with Acknowledgements -- the two may appear in either
% order, but both must be before References), and does not count toward the paper
% page limit. In many cases, where the ethical impacts and expected societal
% implications are those that are well established when advancing the field of
% Machine Learning, substantial discussion is not required, and a simple
% statement such as the following will suffice:

% ``This paper presents work whose goal is to advance the field of Machine
% Learning. There are many potential societal consequences of our work, none
% which we feel must be specifically highlighted here.''

% The above statement can be used verbatim in such cases, but we encourage
% authors to think about whether there is content which does warrant further
% discussion, as this statement will be apparent if the paper is later flagged
% for ethics review.
% }

%\bibliography{example_paper}
\bibliography{references}

\bibliographystyle{ICML2026/icml2026}

%%%%%%%%%%%%%%%%%%%%%%%%%%%%%%%%%%%%%%%%%%%%%%%%%%%%%%%%%%%%%%%%%%%%%%%%%%%%%%%
%%%%%%%%%%%%%%%%%%%%%%%%%%%%%%%%%%%%%%%%%%%%%%%%%%%%%%%%%%%%%%%%%%%%%%%%%%%%%%%
% APPENDIX
%%%%%%%%%%%%%%%%%%%%%%%%%%%%%%%%%%%%%%%%%%%%%%%%%%%%%%%%%%%%%%%%%%%%%%%%%%%%%%%
%%%%%%%%%%%%%%%%%%%%%%%%%%%%%%%%%%%%%%%%%%%%%%%%%%%%%%%%%%%%%%%%%%%%%%%%%%%%%%%
\newpage
\appendix
\onecolumn

\section{Further Background}
\label{sec:further-background}

\paragraph{Bias and Variance of Gradient Estimators.} 
The quality of a gradient estimator $\hat{g}(\theta)$ is characterized by its bias and variance. 
The bias of an estimator quantifies its systematic deviation from the true gradient.
%:
%\begin{equation}
%\text{Bias}[\hat{g}(\theta)] = \mathbb{E}[\hat{g}(\theta)] - \nabla_\theta \mathcal{L}(\theta)$
%\end{equation}
%An estimator is unbiased when $\mathbb{E}[\hat{g}(\theta)] = \nabla_\theta \mathcal{L}(\theta)$, ensuring that it points in the correct optimization direction, on average. 
Biased estimators may converge to suboptimal solutions or fail to converge entirely, due to systematic error accumulation across optimization steps.
The variance of the estimator measures the spread of gradient estimates around their expected value. 
%:
%\begin{equation}
%\text{Var}[\hat{g}(\theta)] = \mathbb{E}[(\hat{g}(\theta) - \mathbb{E}[\hat{g}(\theta)])^2]
%\end{equation}
High variance implies that individual gradient estimates deviate substantially from the true gradient. In such case, more samples are required to obtain reliable updates.
%and potentially necessitating smaller learning rates to maintain training stability.
Because reducing bias often increases variance and vice versa, the bias-variance trade-off represents a fundamental tension in gradient estimator design. 

\paragraph{REINFORCE: An Unbiased but High-Variance Solution.} 
The REINFORCE algorithm~\citep{williams1992simple} provides an unbiased gradient estimator by directly applying the log-derivative trick. Given samples $z^{(1)}, \ldots, z^{(n)} \sim p_\theta(z)$, the REINFORCE estimator computes:
\begin{equation}
\hat{g}_{\text{RF}}(\theta) = \frac{1}{n} \sum_{i=1}^{n} \nabla_\theta \log p_\theta(z^{(i)}) \cdot f(z^{(i)})
\end{equation}
This estimator is provably unbiased, which makes it theoretically attractive, as it guarantees that gradient descent will converge to at least a local optimum given appropriate step sizes and sufficient samples. 
However, REINFORCE suffers from high variance, which, in practice, translates into severe sample inefficiency: obtaining reliable gradient estimates requires averaging over many samples, and optimization proceeds slowly with noisy, unstable updates. 
This inefficiency becomes particularly problematic in reinforcement learning applications, where each sample may require costly environment interactions, and in language model training, where evaluating $f(z)$ for long generated sequences incurs significant computational expense.

\paragraph{Straight-Through Gumbel-Softmax: Bridging Discrete Forward and Continuous Backward.} 
To recover discrete samples during the forward pass while maintaining differentiability in the backward pass,~\citet{bengio2013estimating} introduced the straight-through (ST) estimator. 
The Straight-Through Gumbel-Softmax (STGS) estimator combines this technique with the GS distribution: during the forward pass, a discrete one-hot sample $\mathbf{z} = \text{one\_hot}(\arg\max_k y_k)$ is used, while during the backward pass, gradients flow through the continuous GS relaxation $\mathbf{y}$.
%Formally, the STGS estimator can be understood as using a biased gradient that assumes $\nabla_\theta \mathbf{z} \approx \nabla_\theta \mathbf{y}$. 
This approximation introduces additional bias beyond the standard GS estimator but offers the practical advantage of allowing discrete operations in the forward pass, which may be necessary in some use cases, such as LMs with SoTs as the main abstraction.

\paragraph{The Gap in Quantitative Comparison for Language Models.} 
Despite the theoretical understanding of these gradient estimators and their widespread application in various domains, a comprehensive quantitative comparison of their bias-variance characteristics specifically in the context of LMs remains absent from the literature. 
Given the unique properties of LM tasks---including high-dimensional discrete output spaces (vocabularies of 30,000+ tokens), long-range dependencies in generated sequences, and the coupling between embedding and sampling modules---the relative performance of different gradient estimators may differ substantially from their behavior in other discrete optimization settings. We fill in this gap in the following sections.

%A rigorous empirical characterization of the bias-variance trade-offs would inform principled selection of gradient estimators for different language modeling objectives, enable more effective hyperparameter tuning, and potentially reveal opportunities for domain-specific adaptations of existing methods.

\paragraph{Advanced Gradient Estimators: REBAR and RELAX.} 
Beyond the basic Gumbel-Softmax framework, further developments have sought to address the inherent bias of continuous relaxation methods. The REBAR estimator~\citep{tucker2017rebar} (REinforcement learning with Bootstrapped And Relaxed gradients) combines REINFORCE with a GS-based control variate to achieve an unbiased estimator with reduced variance. By analytically deriving the control variate, REBAR eliminates the bias introduced by the continuous relaxation while maintaining the variance reduction benefits.

Building upon REBAR, the RELAX estimator~\citep{grathwohl2018relax} (REinforcement Learning with eXtra gradient information) further reduces variance by learning a neural network-based control variate that is jointly optimized alongside the main objective. This learned control variate can adapt to the specific structure of the optimization problem, potentially achieving superior variance reduction compared to fixed analytical control variates.

However, unlike the Gumbel-Softmax-based estimators that can be implemented as modular layers within standard automatic differentiation frameworks, REBAR and RELAX require the construction of surrogate loss functions for each discrete variable whose gradients are being estimated. 
This additional scaffolding reduces the plug-and-play nature of these methods, complicating integration into existing codebases and requiring careful implementation to ensure correctness.

Moreover, the introduction of control variates --particularly learned control variates in RELAX-- introduces numerous confounding factors that merit careful investigation. We leave this to future research.
Given these considerations and the desire to maintain methodological clarity in our empirical investigations, we restrict our focus to Gumbel-Softmax-based gradient estimators in this work, leaving comprehensive evaluation of REBAR, RELAX, and related advanced methods to future research.

\pagebreak
\begin{longtable}{|p{2.5cm}|p{2.5cm}|c|p{2cm}|c|p{0.75cm}|p{0.75cm}|p{0.75cm}|}
\caption{Literature review summary table on the LM inversion problem and related works.} \label{table:lit-review-LM-inversion} \\
\hline
\textbf{Paper} & \textbf{Domain} & \textbf{Training-free} & \textbf{LMI Exp.?} & \textbf{Eval. Meth.} & \textbf{Soft Emb.} & \textbf{GS Grad.} & \textbf{Learn $\tau$} \\ \hline
\endfirsthead

\hline
\multicolumn{8}{|c|}{Continuation of Table \ref{table:lit-review-LM-inversion}} \\
\hline
\textbf{Paper} & \textbf{Domain} & \textbf{Training-free} & \textbf{LMI Exp.?} & \textbf{Eval. Meth.} & \textbf{Soft Emb.} & \textbf{GS Grad.} & \textbf{Learn $\tau$} \\ \hline
\endhead

\hline
\endfoot

% --- YOUR DATA START ---
AutoPrompt~\citep{shin-etal-2020-autoprompt-EMNLP2020} & Adversarial attacks & \textcolor{red}{\ding{55}} & \textcolor{red}{\ding{55}} & N/A & N/A & N/A & N/A\\ \hline

Information Leakage~\citep{song2020information} & Embedding inversion & N/A & \textcolor{red}{\ding{55}} & N/A & N/A & N/A & N/A \\ \hline

GBDA~\citep{guo-etal-2021-GBDA-EMNLP2021} & Adversarial attacks & \textcolor{green}{\ding{51}} & \textcolor{red}{\ding{55}} & N/A & \textcolor{green}{\ding{51}} & \textcolor{green}{\ding{51}} & \textcolor{red}{\ding{55}} \\ \hline

ARCA~\citep{jones2023-automatically-auditing-LM-ICML2023} & Adversarial attacks / Auditing & \textcolor{green}{\ding{51}} & \textcolor{green}{\ding{51}} & \parbox{2cm}{\fontsize{6}{6}\selectfont\begin{itemize}[topsep=1pt,partopsep=0pt,leftmargin=*]\item{ARCA} \item{AutoPrompt} \item{GBDA}\end{itemize}} & N/A & N/A & N/A \\ \hline

GCG~\citep{zou2023GCG} & Adversarial attacks & \textcolor{green}{\ding{51}} & \parbox{2cm}{Fluent\\ Harmful\\ Output Elicit.} & \parbox{2cm}{\fontsize{6}{6}\selectfont\begin{itemize}[topsep=1pt,partopsep=0pt,leftmargin=*]\item{GCG} \item{PEZ} \item{AutoPrompt} \item{GBDA} \end{itemize}} & N/A & N/A & N/A \\ \hline

PEZ~\citep{wen2023hard-PEZ} & Adversarial attacks & \textcolor{green}{\ding{51}} & Image Output Elicit. & \parbox{2cm}{\fontsize{6}{6}\selectfont\begin{itemize}[topsep=1pt,partopsep=0pt,leftmargin=*]\item{PEZ} \item{AutoPrompt} \item{FluentPrompt} \end{itemize}} & N/A & N/A & N/A \\ \hline

Vec2Text~\citep{morris2023Vec2text} & Embedding Inversion & \textcolor{red}{\ding{55}} & \textcolor{red}{\ding{55}} & N/A & N/A & N/A & N/A \\ \hline

Logit-to-Text (L2T-\citet{morris2024LMI}) & (Fluent-only) LMI & \textcolor{red}{\ding{55}} & Prompt Rec. \textbf{from logits} & \parbox{2cm}{\fontsize{6}{6}\selectfont\begin{itemize}[topsep=1pt,partopsep=0pt,leftmargin=*]\item{L2T} \item{Few-Shot LLM}\end{itemize}} & N/A & N/A & N/A \\ \hline

output2prompt (O2P-\citet{zhang-etal-2024-extracting-O2P}) & (Fluent-only) LMI & \textcolor{red}{\ding{55}} & Prompt Rec. \textbf{from text} & \parbox{2cm}{\fontsize{6}{6}\selectfont\begin{itemize}[topsep=1pt,partopsep=0pt,leftmargin=*]\item{O2P} \item{L2T} \item{Few-Shot LLM}\end{itemize}} & N/A & N/A & N/A \\ \hline

PILS~\citep{nazir2025better-PILS} & (Fluent-only) LMI & \textcolor{red}{\ding{55}} & Prompt Rec. \textbf{from logits} & \parbox{2cm}{\fontsize{6}{6}\selectfont\begin{itemize}[topsep=1pt,partopsep=0pt,leftmargin=*]\item{PILS} \item{L2T} \item{L2T++} \item{O2P}\end{itemize}} & N/A & N/A & N/A \\ \hline

Reverse Prompt Engineering (RPE-\citet{li-klabjan-2025-RPE}) & (Fluent-only) LMI & \textcolor{green}{\ding{51}} & Prompt Rec. \textbf{from logits or text} & \parbox{2cm}{\fontsize{6}{6}\selectfont\begin{itemize}[topsep=1pt,partopsep=0pt,leftmargin=*]\item{RPE} \item{RPEGA} \item{O2P} \item{O2Ps}\end{itemize}} & N/A & N/A & N/A \\ \hline

SODA~\citep{skapars2025SODA-gpt} & (Fluent \& \textbf{Non-Fluent}) LMI & \textcolor{red}{\ding{55}} & Prompt Rec. \textbf{from logits or text} & \parbox{2cm}{\fontsize{6}{6}\selectfont\begin{itemize}[topsep=1pt,partopsep=0pt,leftmargin=*]\item{SODA} \item{GCG} \item{L2T}\end{itemize}} & \textcolor{green}{\ding{51}} & \textcolor{red}{\ding{55}} & \textcolor{red}{\ding{55}} \\ \hline

\textbf{Ours} & \textbf{Fluency-Parameterised} LMI & \textcolor{green}{\ding{51}} & \textcolor{green}{\ding{51}} & \parbox{2cm}{\fontsize{6}{6}\selectfont\begin{itemize}[topsep=1pt,partopsep=0pt,leftmargin=*]\item{Ours} \item{GBDA} \item{REINFORCE} \item{SODA} \end{itemize}} & \textcolor{green}{\ding{51}} & \textcolor{green}{\ding{51}} & \textcolor{green}{\ding{51}} \\ \hline
% --- YOUR DATA END ---

\end{longtable}

\pagebreak
\section{Further Qualitative Results}
\label{sec:further-qualitative-results}

The following tables~\ref{tab:optimization_results_s2} and ~\ref{tab:optimization_results_s3} show qualitative results of our \textbf{DLMI($\tau_0$$=$$100$)} for $N=10$ and $M=20$.

% Requires: \usepackage{graphicx}, \usepackage{xcolor}
% Image directory: table_images_s2/

\begin{longtable}{c|c|c|c|c|c|c}
\caption{Qualitative optimization results for SmolLM2-135M ; PPL = Perplexity ;{\footnotesize\color[rgb]{0.55,0.55,0.55} \texttt{\textbackslash n} = newline, \texttt{\textbackslash t} = tabulation, \textvisiblespace{} = whitespace.} {\footnotesize\color[rgb]{0.78,0.47,0.47} Reddish text in Sampled Output cells corresponds to differences from Target Output cells.}} \label{tab:optimization_results_s2} \\
\hline
\textbf{ID} & \textbf{Rank $k$} & \textbf{Target Output} & \textbf{Learned Prompt} & \textbf{Prompt PPL} & \textbf{Sampled Output} & \textbf{LCS} \\
\hline
\endfirsthead

\multicolumn{7}{c}{{\bfseries \tablename\ \thetable{} -- continued from previous page}} \\
\hline
\textbf{ID} & \textbf{Rank $k$} & \textbf{Target Output} & \textbf{Learned Prompt} & \textbf{Prompt PPL} & \textbf{Sampled Output} & \textbf{LCS} \\
\hline
\endhead

\hline \multicolumn{7}{|r|}{{Continued on next page}} \\
\hline
\endfoot

\hline
\endlastfoot

k1\_sample0 & 1 & \includegraphics[width=2.0cm]{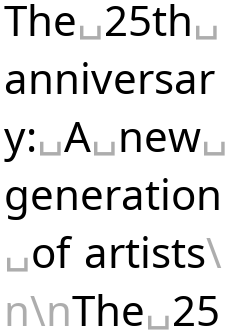} & \includegraphics[width=2.0cm]{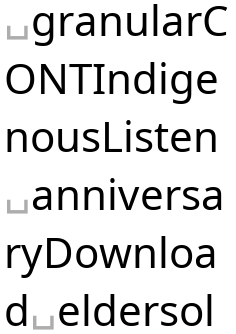} & 6906479.50 & \includegraphics[width=2.0cm]{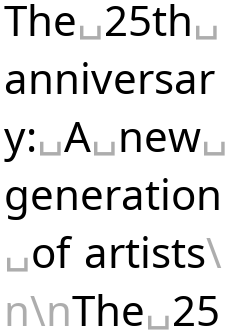} & 1.000 \\ \hline
k1\_sample1 & 1 & \includegraphics[width=2.0cm]{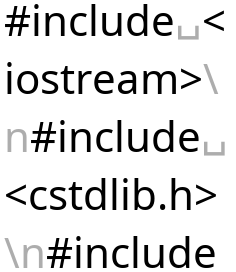} & \includegraphics[width=2.0cm]{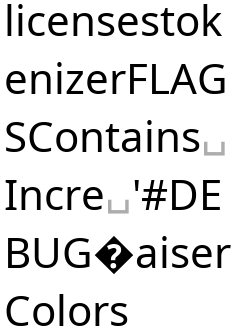} & 2976247.00 & \includegraphics[width=2.0cm]{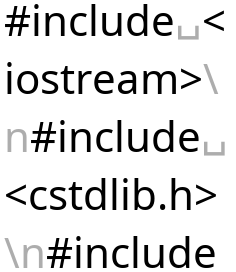} & 1.000 \\ \hline
k1\_sample2 & 1 & \includegraphics[width=2.0cm]{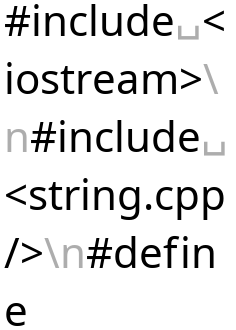} & \includegraphics[width=2.0cm]{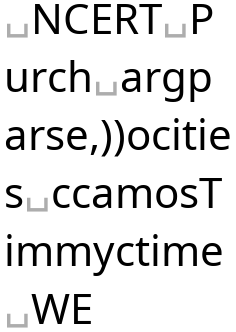} & 1490148.00 & \includegraphics[width=2.0cm]{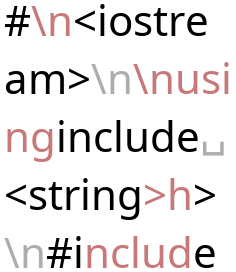} & 0.600 \\ \hline
k1\_sample3 & 1 & \includegraphics[width=2.0cm]{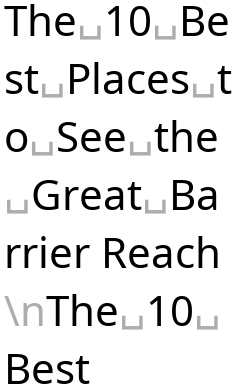} & \includegraphics[width=2.0cm]{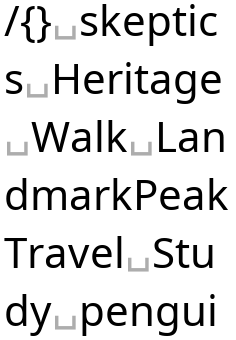} & 501437.19 & \includegraphics[width=2.0cm]{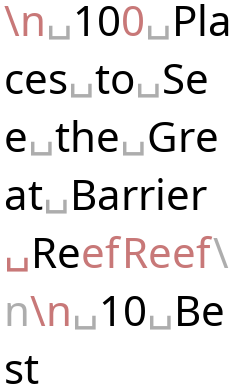} & 0.750 \\ \hline
k1\_sample4 & 1 & \includegraphics[width=2.0cm]{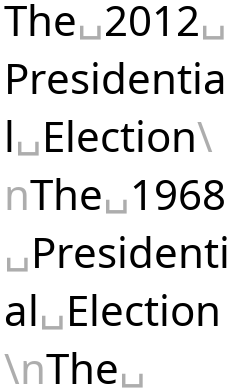} & \includegraphics[width=2.0cm]{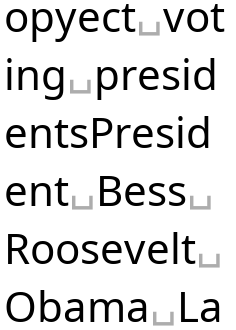} & 188419.69 & \includegraphics[width=2.0cm]{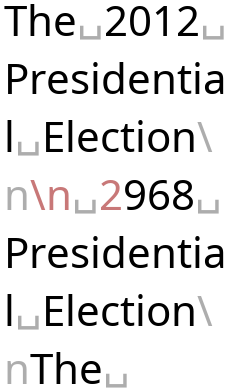} & 0.900 \\ \hline
k6\_sample0 & 6 & \includegraphics[width=2.0cm]{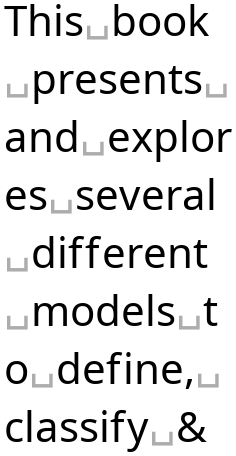} & \includegraphics[width=2.0cm]{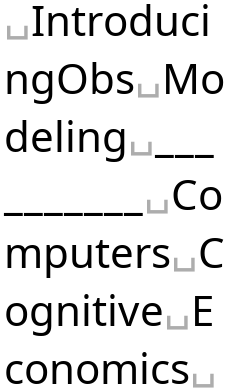} & 337010.06 & \includegraphics[width=2.0cm]{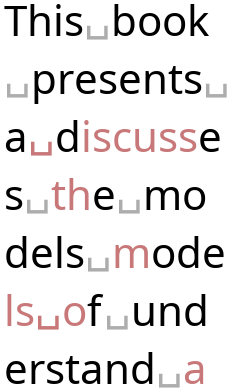} & 0.350 \\ \hline
k6\_sample1 & 6 & \includegraphics[width=2.0cm]{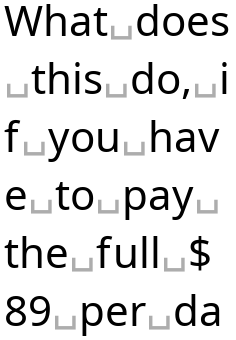} & \includegraphics[width=2.0cm]{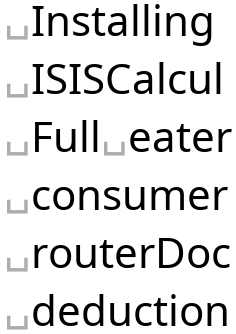} & 2646392.00 & \includegraphics[width=2.0cm]{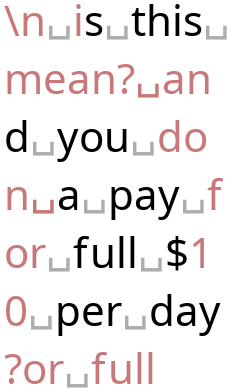} & 0.350 \\ \hline
k6\_sample2 & 6 & \includegraphics[width=2.0cm]{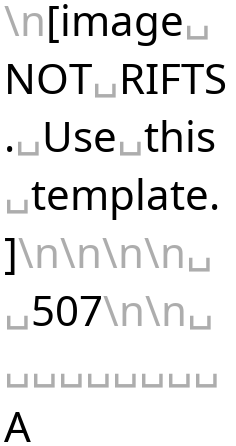} & \includegraphics[width=2.0cm]{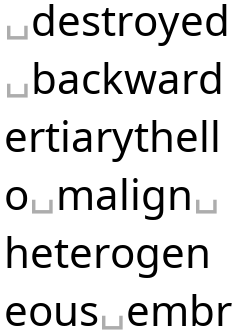} & 1077817.38 & \includegraphics[width=2.0cm]{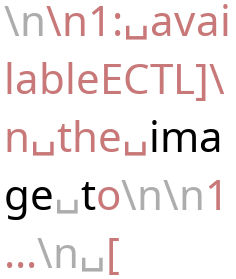} & 0.150 \\ \hline
k6\_sample3 & 6 & \includegraphics[width=2.0cm]{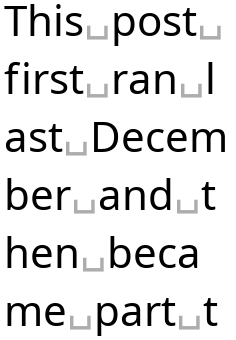} & \includegraphics[width=2.0cm]{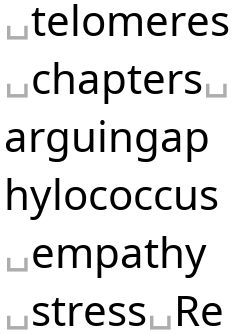} & 1883703.75 & \includegraphics[width=2.0cm]{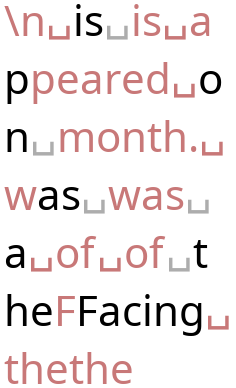} & 0.200 \\ \hline
k6\_sample4 & 6 & \includegraphics[width=2.0cm]{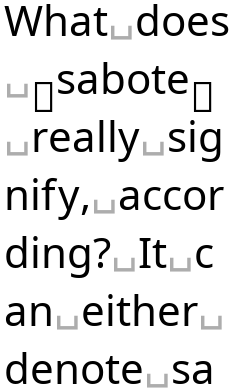} & \includegraphics[width=2.0cm]{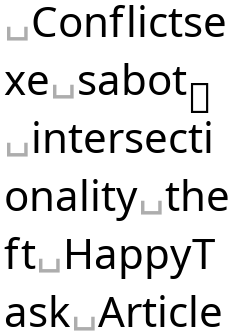} & 6997480.50 & \includegraphics[width=2.0cm]{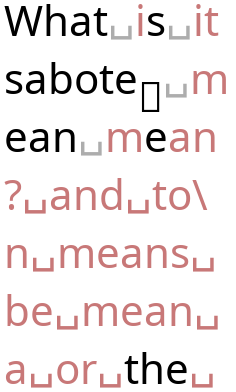} & 0.300 \\ \hline
k11\_sample0 & 11 & \includegraphics[width=2.0cm]{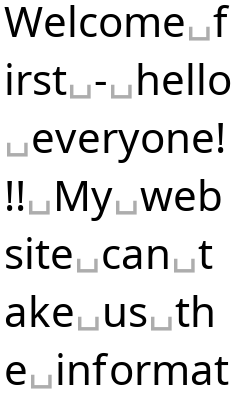} & \includegraphics[width=2.0cm]{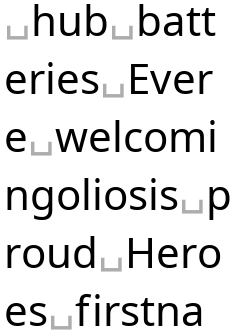} & 786451.88 & \includegraphics[width=2.0cm]{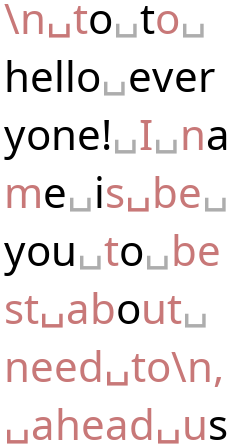} & 0.200 \\ \hline
k11\_sample1 & 11 & \includegraphics[width=2.0cm]{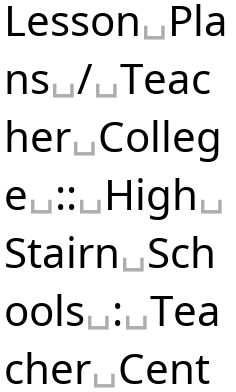} & \includegraphics[width=2.0cm]{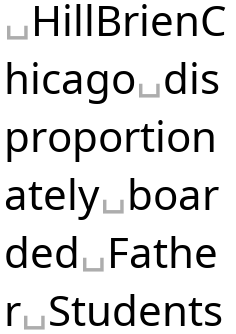} & 833458.31 & \includegraphics[width=2.0cm]{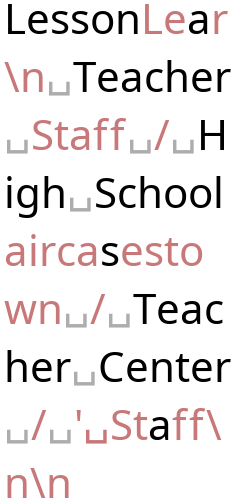} & 0.300 \\ \hline
k11\_sample2 & 11 & \includegraphics[width=2.0cm]{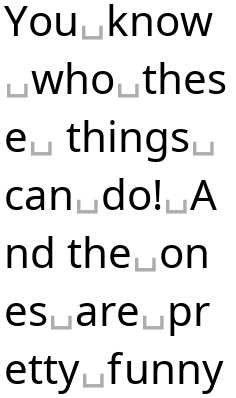} & \includegraphics[width=2.0cm]{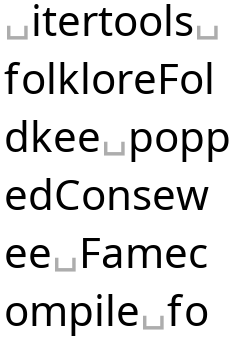} & 840190.62 & \includegraphics[width=2.0cm]{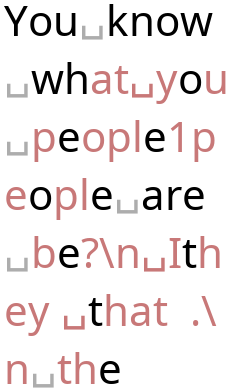} & 0.250 \\ \hline
k11\_sample3 & 11 & \includegraphics[width=2.0cm]{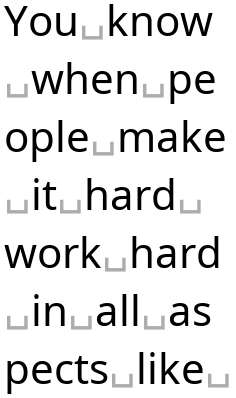} & \includegraphics[width=2.0cm]{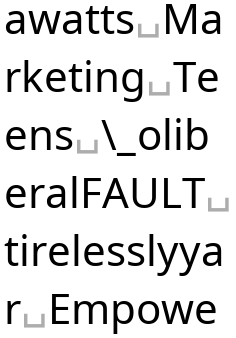} & 2217176.00 & \includegraphics[width=2.0cm]{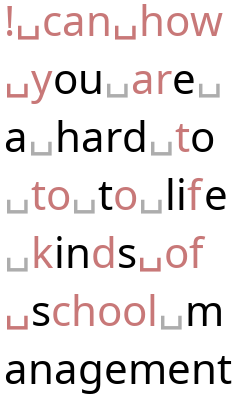} & 0.150 \\ \hline
k11\_sample4 & 11 & \includegraphics[width=2.0cm]{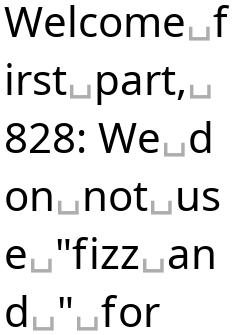} & \includegraphics[width=2.0cm]{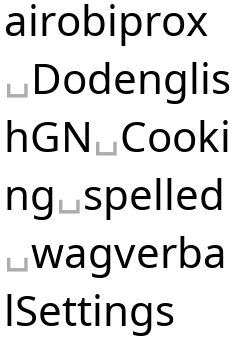} & 1620452.25 & \includegraphics[width=2.0cm]{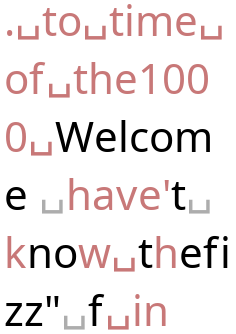} & 0.150 \\ \hline
k16\_sample0 & 16 & \includegraphics[width=2.0cm]{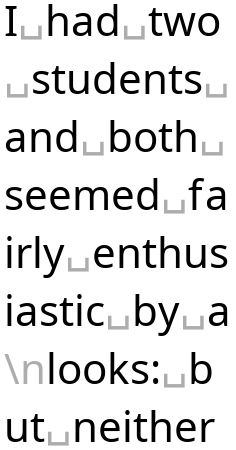} & \includegraphics[width=2.0cm]{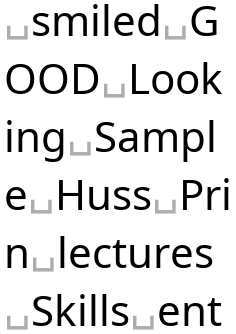} & 1746329.75 & \includegraphics[width=2.0cm]{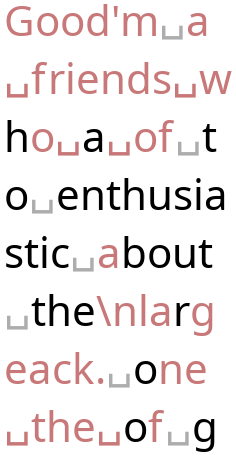} & 0.150 \\ \hline
k16\_sample1 & 16 & \includegraphics[width=2.0cm]{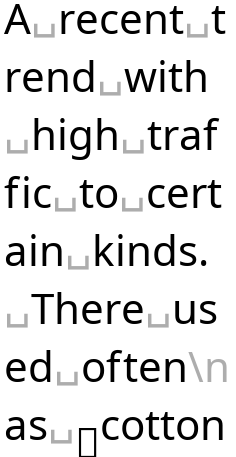} & \includegraphics[width=2.0cm]{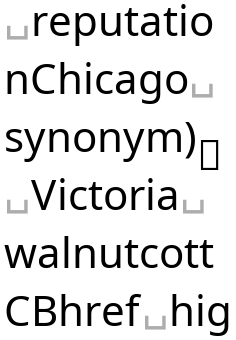} & 2399038.25 & \includegraphics[width=2.0cm]{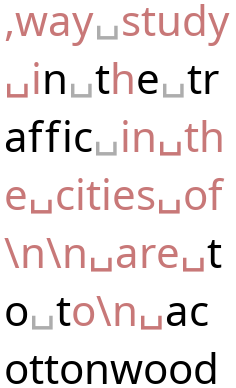} & 0.300 \\ \hline
k16\_sample2 & 16 & \includegraphics[width=2.0cm]{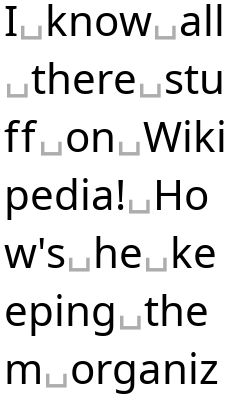} & \includegraphics[width=2.0cm]{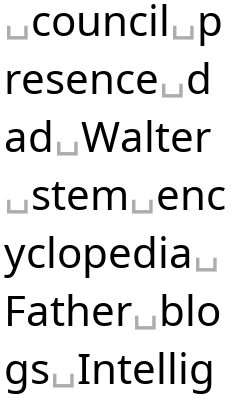} & 264530.31 & \includegraphics[width=2.0cm]{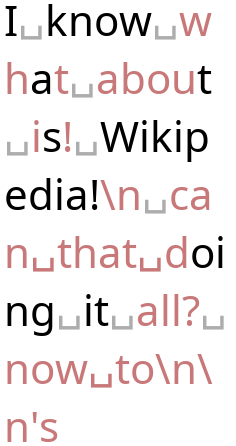} & 0.250 \\ \hline
k16\_sample3 & 16 & \includegraphics[width=2.0cm]{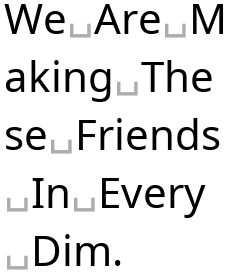} & \includegraphics[width=2.0cm]{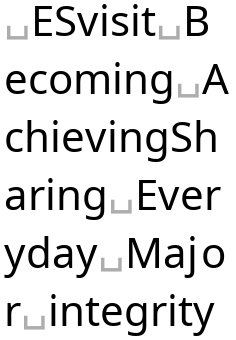} & 263368.09 & \includegraphics[width=2.0cm]{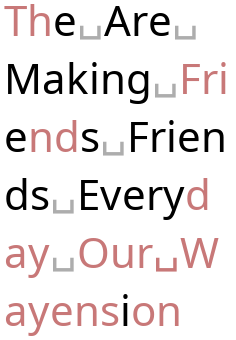} & 0.333 \\ \hline
k16\_sample4 & 16 & \includegraphics[width=2.0cm]{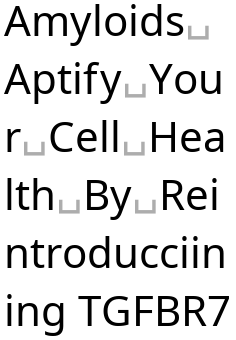} & \includegraphics[width=2.0cm]{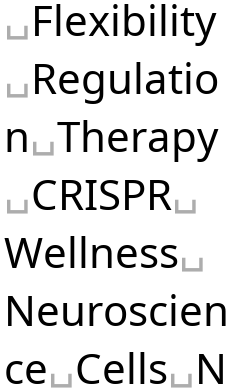} & 147764.72 & \includegraphics[width=2.0cm]{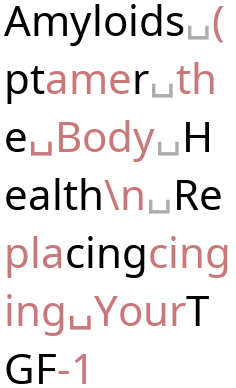} & 0.450 \\ \hline
k21\_sample0 & 21 & \includegraphics[width=2.0cm]{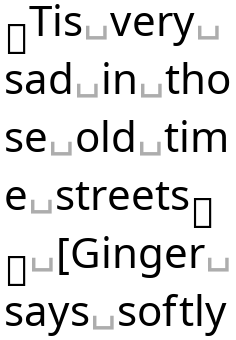} & \includegraphics[width=2.0cm]{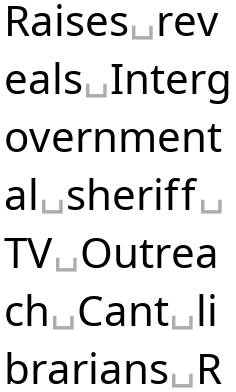} & 1902028.00 & \includegraphics[width=2.0cm]{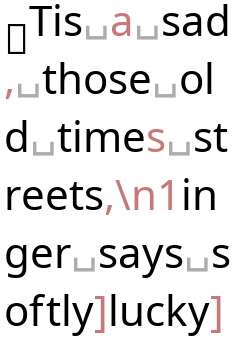} & 0.600 \\ \hline
k21\_sample1 & 21 & \includegraphics[width=2.0cm]{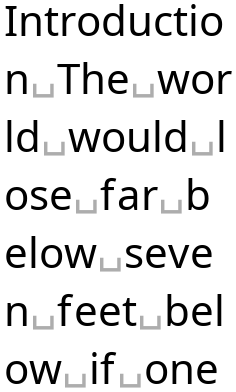} & \includegraphics[width=2.0cm]{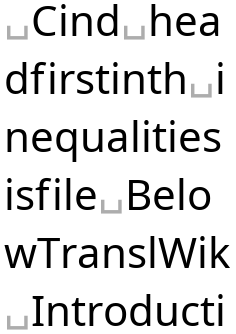} & 4995487.50 & \includegraphics[width=2.0cm]{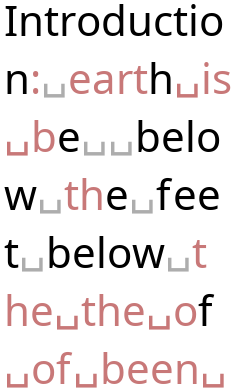} & 0.250 \\ \hline
k21\_sample2 & 21 & \includegraphics[width=2.0cm]{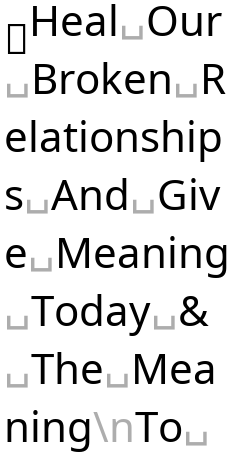} & \includegraphics[width=2.0cm]{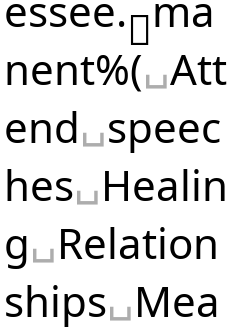} & 319092.66 & \includegraphics[width=2.0cm]{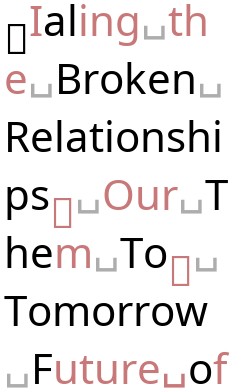} & 0.250 \\ \hline
k21\_sample3 & 21 & \includegraphics[width=2.0cm]{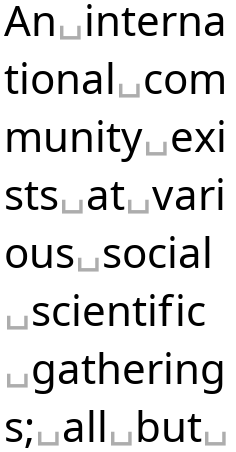} & \includegraphics[width=2.0cm]{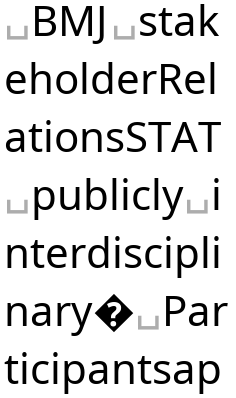} & 4426985.00 & \includegraphics[width=2.0cm]{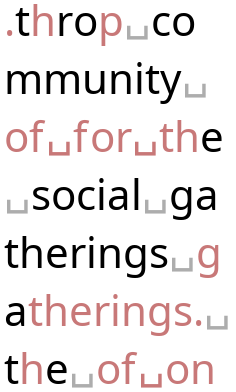} & 0.250 \\ \hline
k21\_sample4 & 21 & \includegraphics[width=2.0cm]{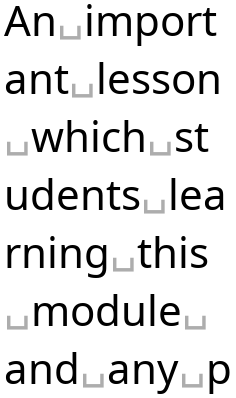} & \includegraphics[width=2.0cm]{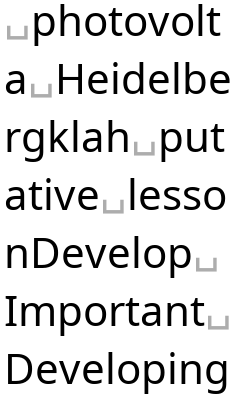} & 350061.91 & \includegraphics[width=2.0cm]{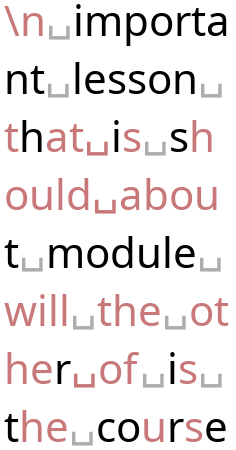} & 0.300 \\ \hline
\end{longtable}
% Requires: \usepackage{graphicx}, \usepackage{xcolor}
% Image directory: table_images_s3/

\begin{longtable}{c|c|c|c|c|c|c}
\caption{Qualitative optimization results for SmolLM3-base-3B ; PPL = Perplexity ; {\footnotesize\color[rgb]{0.55,0.55,0.55} \texttt{\textbackslash n} = newline, \texttt{\textbackslash t} = tabulation, \textvisiblespace{} = whitespace.} {\footnotesize\color[rgb]{0.78,0.47,0.47} Reddish text in Sampled Output cells corresponds to differences from Target Output cells.}}  \\
\label{tab:optimization_results_s3} \\
\hline
\textbf{ID} & \textbf{Rank $k$} & \textbf{Target Output} & \textbf{Learned Prompt} & \textbf{Prompt PPL} & \textbf{Sampled Output} & \textbf{LCS} \\
\hline
\endfirsthead

\multicolumn{7}{c}{{\bfseries \tablename\ \thetable{} -- continued from previous page}} \\
\hline
\textbf{ID} & \textbf{Rank $k$} & \textbf{Target Output} & \textbf{Learned Prompt} & \textbf{Prompt PPL} & \textbf{Sampled Output} & \textbf{LCS} \\
\hline
\endhead

\hline \multicolumn{7}{|r|}{{Continued on next page}} \\
\hline
\endfoot

\hline
\endlastfoot

k1\_sample0 & 1 & \includegraphics[width=2.0cm]{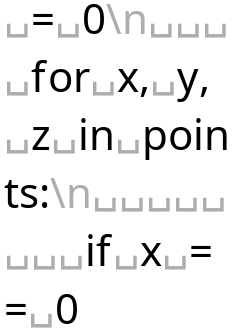} & \includegraphics[width=2.0cm]{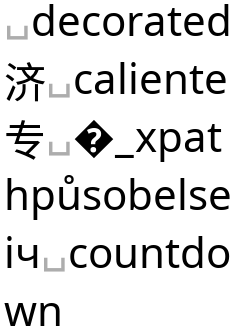} & 11062909.00 & \includegraphics[width=2.0cm]{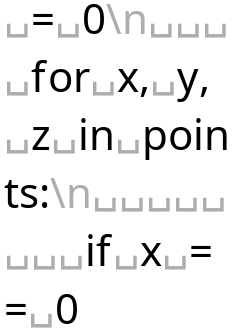} & 1.000 \\ \hline
k1\_sample1 & 1 & \includegraphics[width=2.0cm]{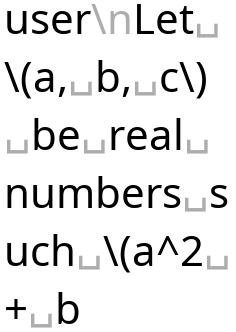} & \includegraphics[width=2.0cm]{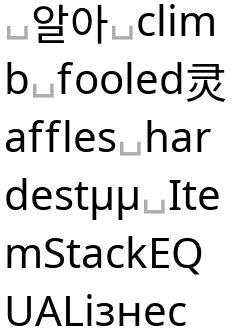} & 12334413.00 & \includegraphics[width=2.0cm]{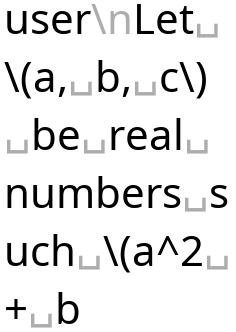} & 1.000 \\ \hline
k1\_sample2 & 1 & \includegraphics[width=2.0cm]{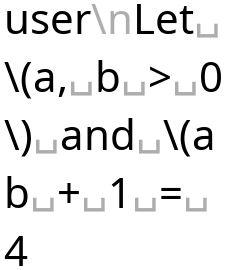} & \includegraphics[width=2.0cm]{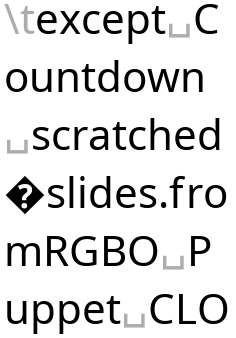} & 11888118.00 & \includegraphics[width=2.0cm]{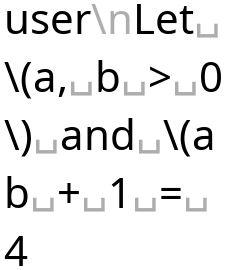} & 1.000 \\ \hline
k1\_sample3 & 1 & \includegraphics[width=2.0cm]{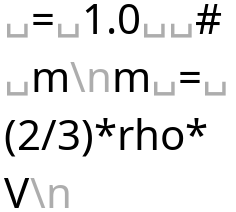} & \includegraphics[width=2.0cm]{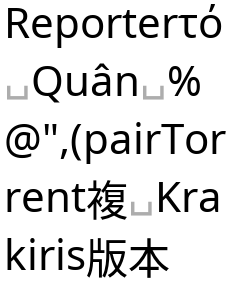} & 11281364.00 & \includegraphics[width=2.0cm]{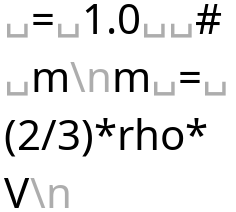} & 1.000 \\ \hline
k1\_sample4 & 1 & \includegraphics[width=2.0cm]{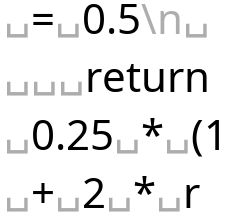} & \includegraphics[width=2.0cm]{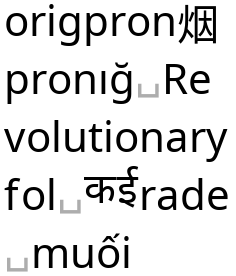} & 8558886.00 & \includegraphics[width=2.0cm]{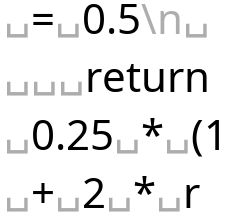} & 1.000 \\ \hline
k6\_sample0 & 6 & \includegraphics[width=2.0cm]{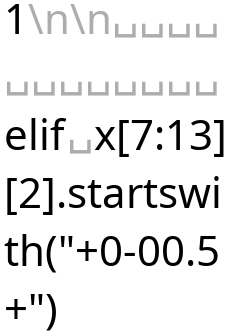} & \includegraphics[width=2.0cm]{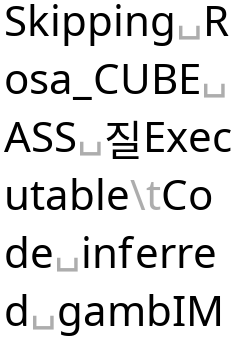} & 9863758.00 & \includegraphics[width=2.0cm]{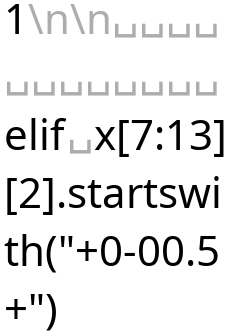} & 1.000 \\ \hline
k6\_sample1 & 6 & \includegraphics[width=2.0cm]{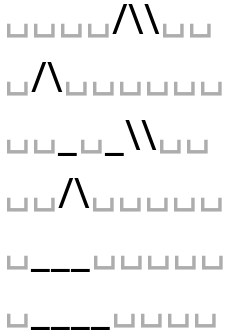} & \includegraphics[width=2.0cm]{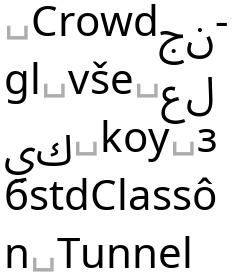} & 11695687.00 & \includegraphics[width=2.0cm]{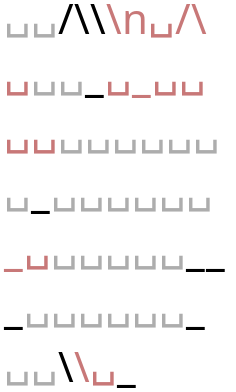} & 0.450 \\ \hline
k6\_sample2 & 6 & \includegraphics[width=2.0cm]{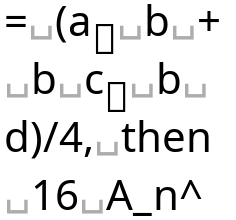} & \includegraphics[width=2.0cm]{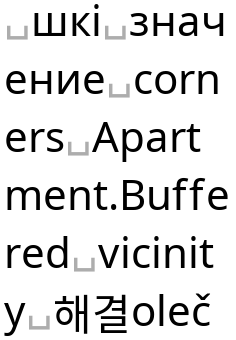} & 1733220.62 & \includegraphics[width=2.0cm]{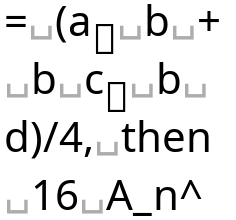} & 1.000 \\ \hline
k6\_sample3 & 6 & \includegraphics[width=2.0cm]{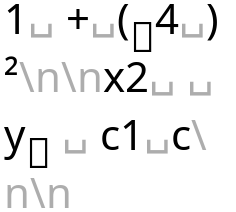} & \includegraphics[width=2.0cm]{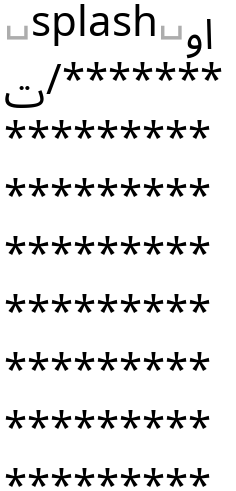} & 25233958.00 & \includegraphics[width=2.0cm]{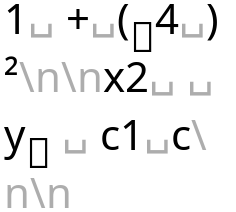} & 1.000 \\ \hline
k6\_sample4 & 6 & \includegraphics[width=2.0cm]{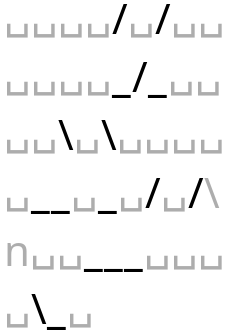} & \includegraphics[width=2.0cm]{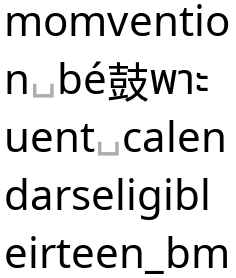} & 7958601.50 & \includegraphics[width=2.0cm]{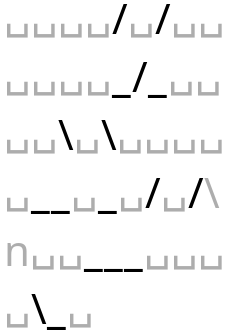} & 1.000 \\ \hline
k11\_sample0 & 11 & \includegraphics[width=2.0cm]{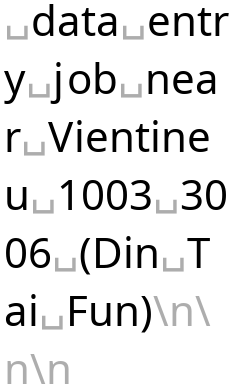} & \includegraphics[width=2.0cm]{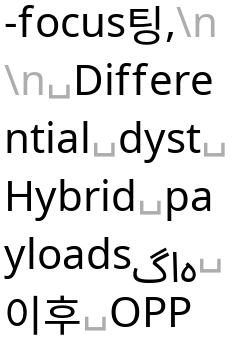} & 5733929.00 & \includegraphics[width=2.0cm]{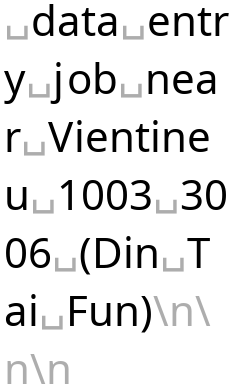} & 1.000 \\ \hline
k11\_sample1 & 11 & \includegraphics[width=2.0cm]{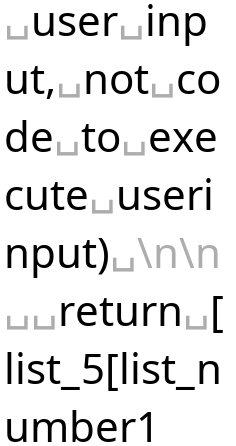} & \includegraphics[width=2.0cm]{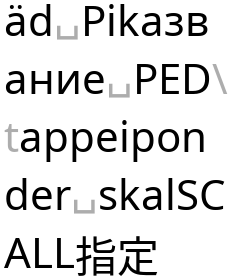} & 14313984.00 & \includegraphics[width=2.0cm]{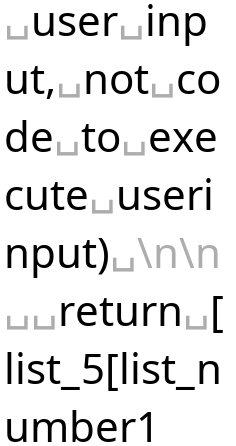} & 1.000 \\ \hline
k11\_sample2 & 11 & \includegraphics[width=2.0cm]{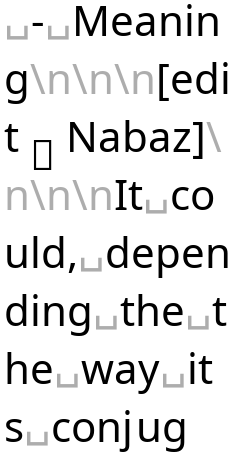} & \includegraphics[width=2.0cm]{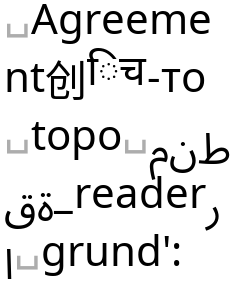} & 3690836.25 & \includegraphics[width=2.0cm]{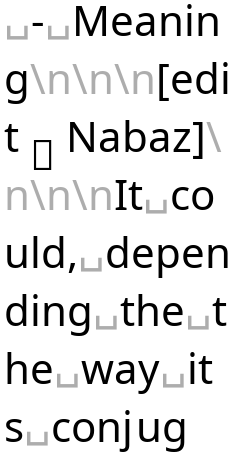} & 1.000 \\ \hline
k11\_sample3 & 11 & \includegraphics[width=2.0cm]{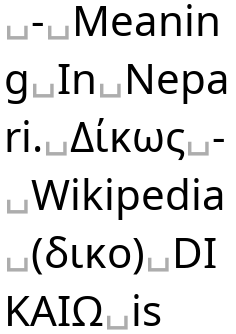} & \includegraphics[width=2.0cm]{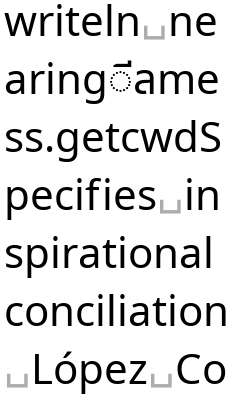} & 23371182.00 & \includegraphics[width=2.0cm]{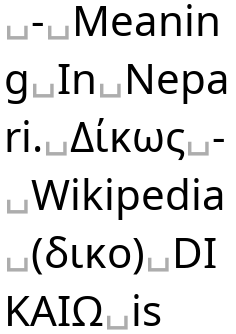} & 1.000 \\ \hline
k11\_sample4 & 11 & \includegraphics[width=2.0cm]{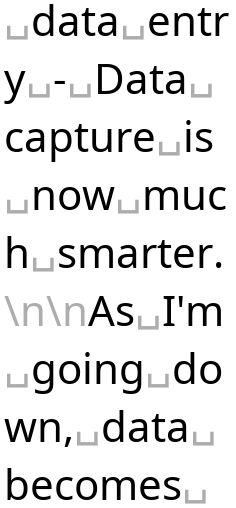} & \includegraphics[width=2.0cm]{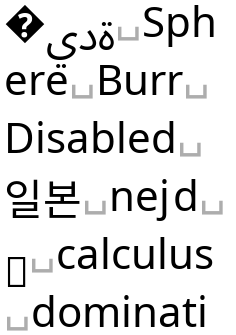} & 2976485.50 & \includegraphics[width=2.0cm]{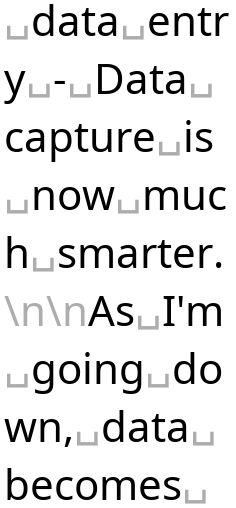} & 1.000 \\ \hline
k16\_sample0 & 16 & \includegraphics[width=2.0cm]{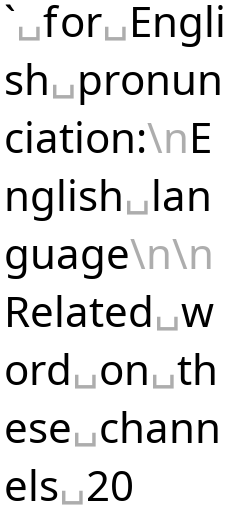} & \includegraphics[width=2.0cm]{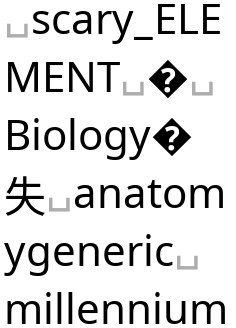} & 10783535.00 & \includegraphics[width=2.0cm]{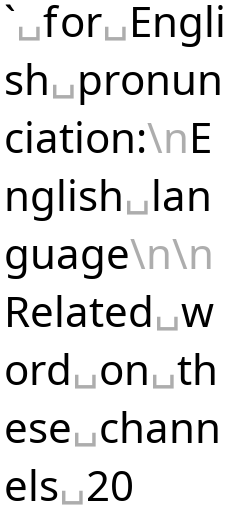} & 1.000 \\ \hline
k16\_sample1 & 16 & \includegraphics[width=2.0cm]{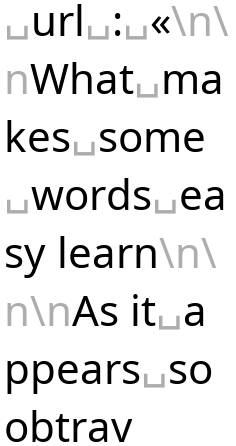} & \includegraphics[width=2.0cm]{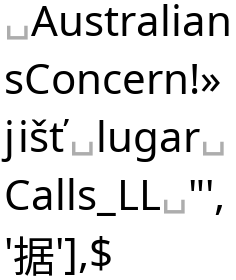} & 10553383.00 & \includegraphics[width=2.0cm]{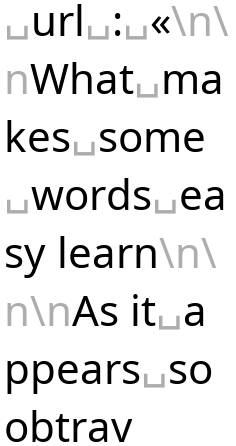} & 1.000 \\ \hline
k16\_sample2 & 16 & \includegraphics[width=2.0cm]{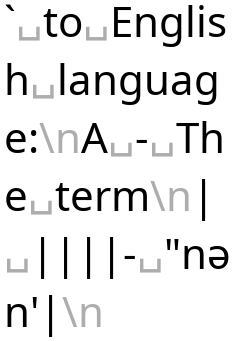} & \includegraphics[width=2.0cm]{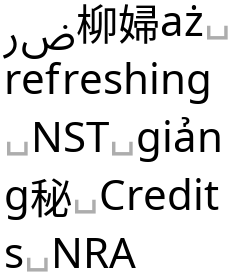} & 6669902.00 & \includegraphics[width=2.0cm]{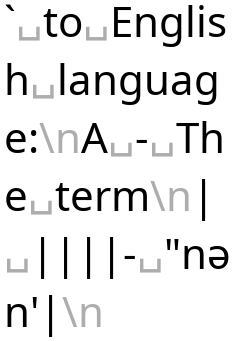} & 1.000 \\ \hline
k16\_sample3 & 16 & \includegraphics[width=2.0cm]{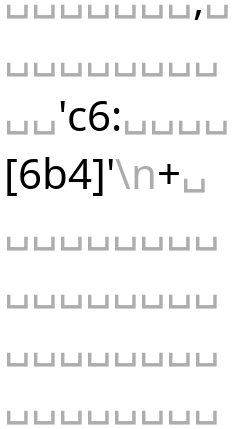} & \includegraphics[width=2.0cm]{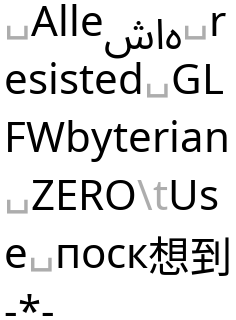} & 18383480.00 & \includegraphics[width=2.0cm]{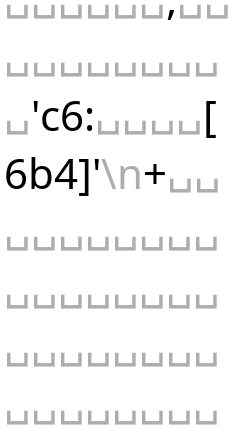} & 1.000 \\ \hline
k16\_sample4 & 16 & \includegraphics[width=2.0cm]{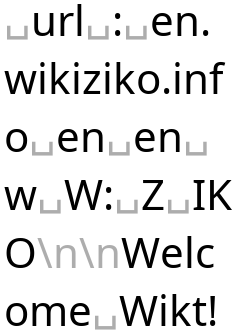} & \includegraphics[width=2.0cm]{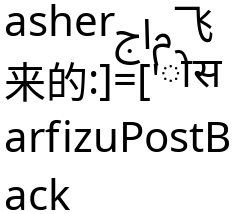} & 10071568.00 & \includegraphics[width=2.0cm]{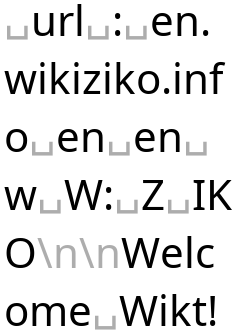} & 1.000 \\ \hline
k21\_sample0 & 21 & \includegraphics[width=2.0cm]{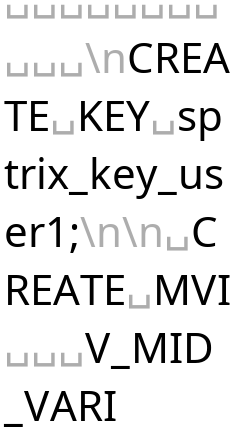} & \includegraphics[width=2.0cm]{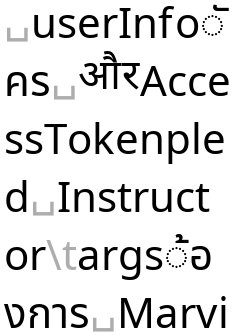} & 23757254.00 & \includegraphics[width=2.0cm]{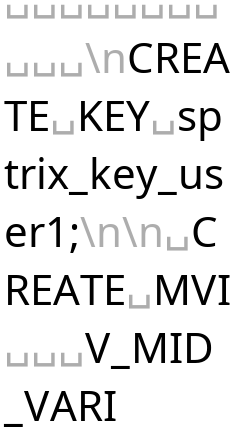} & 1.000 \\ \hline
k21\_sample1 & 21 & \includegraphics[width=2.0cm]{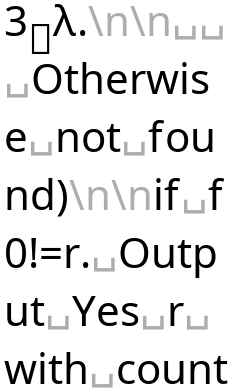} & \includegraphics[width=2.0cm]{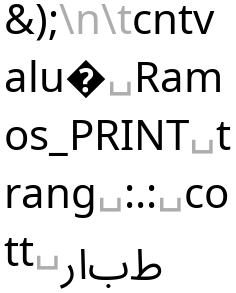} & 30721778.00 & \includegraphics[width=2.0cm]{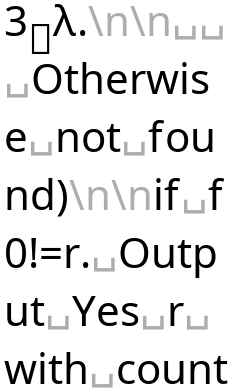} & 1.000 \\ \hline
k21\_sample2 & 21 & \includegraphics[width=2.0cm]{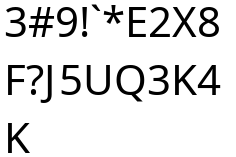} & \includegraphics[width=2.0cm]{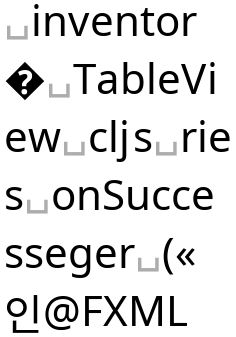} & 5704171.50 & \includegraphics[width=2.0cm]{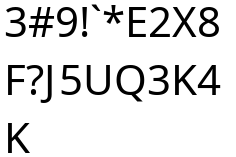} & 1.000 \\ \hline
k21\_sample3 & 21 & \includegraphics[width=2.0cm]{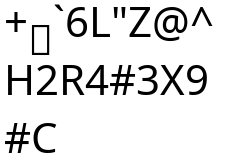} & \includegraphics[width=2.0cm]{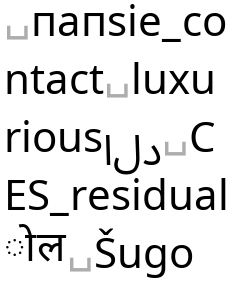} & 25182656.00 & \includegraphics[width=2.0cm]{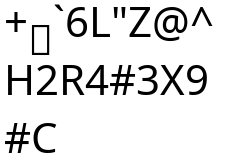} & 1.000 \\ \hline
k21\_sample4 & 21 & \includegraphics[width=2.0cm]{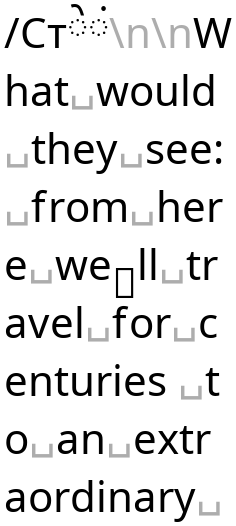} & \includegraphics[width=2.0cm]{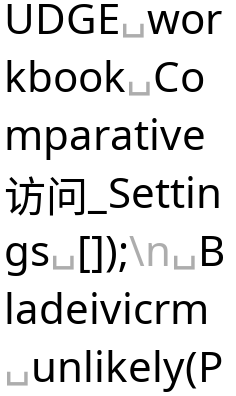} & 14826782.00 & \includegraphics[width=2.0cm]{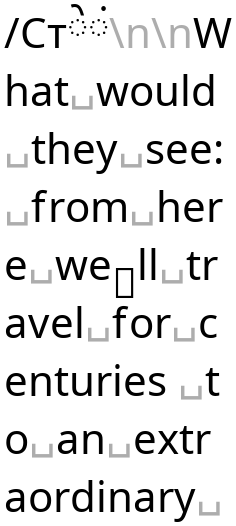} & 1.000 \\ \hline
\end{longtable}

% \begin{table}[ht]
%     \centering
%     \caption{Experimental results showing optimized prompts and sampled outputs.}
%     \label{tab:results}
%     % Adjust width to \textwidth to make it fit the page perfectly
%     \includegraphics[width=\textwidth]{tables/table_pdf.pdf}
% \end{table}

\pagebreak
\section{Hyperparameters}
\label{sec:hyperparams}

Table~\ref{tab:hyperparams} present the hyperparameters with which we ran the experiments presented in Section~\ref{sec:experiments}.

\begin{table}[ht]
\centering
\caption{Hyperparameter configurations for REINFORCE, SODA, and DLMI variants.}
\label{tab:hyperparams}
\begin{adjustbox}{width=\textwidth}
\begin{tabular}{@{}llllll@{}}
\toprule
\textbf{Hyperparameter} & \textbf{Description} & \textbf{REINFORCE} & \textbf{SODA} & \textbf{DLMI($\tau_0$$=$$100$)-TF} & \textbf{DLMI($\tau_0$$=$$100$)} \\ \midrule
%batch\_size & Optimization batch size & 8 & 8 & 8 & 8 \\
num\_samples & Number of samples for grad. est. & 8 & N/A & 8 & 8 \\
decouple\_learnable\_temp & Learn multiple temp. parameters & N/A & N/A & true & true \\
epochs & Number of optimization epochs & [256,2048] & [256,2048] & [256,2048] & [256,2048] \\
gradient\_estimator & Estimator for discrete inputs & reinforce & N/A & gs & gs \\
%learnable\_temperature & Learn temperature parameter & N/A & N/A & true & true \\
learning\_rate & Learning rate for optimization & 0.1 & 0.03 & 0.1 & 0.1 \\
%model\_name & Language model name & \multicolumn{4}{c}{HuggingFaceTB/SmolLM3-3B-Base} \\
reinforce\_baseline\_beta & EMA coefficient for baseline & 0.9 & N/A & N/A & N/A \\
reinforce\_reward\_scale & Policy loss scaling factor & 1 & N/A & N/A & N/A \\
reinforce\_use\_baseline & Enable variance reduction & true & N/A & N/A & N/A \\
seq\_len & Sequence length to optimize & [10,80] & [10,80] & [10,80] & [10,80] \\
soda\_beta1 & SODA: Adam beta1 & N/A & 0.9 & N/A & N/A \\
soda\_beta2 & SODA: Adam beta2 & N/A & 0.995 & N/A & N/A \\
soda\_bias\_correction & SODA: Adam bias correction & N/A & false & N/A & N/A \\
soda\_decay\_rate & Embedding decay rate & N/A & 0.98 & N/A & N/A \\
soda\_reinit\_epoch & Reinit frequency & N/A & 1500 & N/A & N/A \\
soda\_reset\_epoch & Optimizer reset frequency & N/A & 50 & N/A & N/A \\
init\_strategy & Embedding init strategy & N/A & Zeros & Gaussian & Gaussian \\
%stgs\_hard & Use hard ST-GS & N/A & N/A & false & false \\
teacher\_forcing & Use teacher forcing & N/A & N/A & false & true \\
temperature & Fixed or initial temperature $\tau$/$\tau_0$ & N/A & 0.05 & 100 & 100 \\ \bottomrule
\end{tabular}
\end{adjustbox}
\end{table}

\end{document}